\documentclass[letterpaper]{article} 
\usepackage{aaai2026}
\usepackage{times}  
\usepackage{helvet}  
\usepackage{rotating}
\usepackage{makecell}
\usepackage{tcolorbox}
\usepackage{cuted} 
\usepackage{courier}  
\usepackage[hyphens]{url}  
\usepackage{booktabs}   
\usepackage{siunitx}    
\usepackage{multirow}   
\usepackage{graphicx} 

\usepackage{booktabs}   
\usepackage{siunitx}    
\usepackage{multirow}   
\usepackage{graphicx}   

\usepackage{booktabs}   
\usepackage{multirow}   
\usepackage{siunitx}    

\usepackage{pifont}       
\usepackage{bbding}       
\usepackage{fontawesome} 
\definecolor{mygreen}{HTML}{1E8449} 
\definecolor{myred}{HTML}{C0392B}  
\usepackage{xcolor}     
\usepackage{colortbl}  
\usepackage{natbib}  
\usepackage{caption} 
\usepackage{booktabs}
\usepackage{algorithm}
\usepackage{algorithmic}

\usepackage{newfloat}
\usepackage{listings}
\DeclareCaptionStyle{ruled}{labelfont=normalfont,labelsep=colon,strut=off} 
\floatstyle{ruled}
\newfloat{listing}{tb}{lst}{}
\floatname{listing}{Listing}
\usepackage{amsfonts} 
\usepackage{amsmath}
\usepackage{booktabs}
\usepackage{siunitx}
\usepackage{multirow}

\usepackage{tcolorbox}

\nocopyright 

\title{A Rolling Stone Gathers No Moss: Adaptive Policy Optimization for Stable Self-Evaluation in Large
Multimodal Models}
\author{
    Wenkai Wang\textsuperscript{\rm 1},
    Hongcan Guo\textsuperscript{\rm 2},
    Zheqi Lv\textsuperscript{\rm 1},
    Shengyu Zhang\textsuperscript{\rm 1}
}
\affiliations{
    \textsuperscript{\rm 1} Zhejiang University, Hangzhou, China\\
    \textsuperscript{\rm 2} Beijing University of Posts and Telecommunications, China\\

}

\usepackage{bibentry}

\begin{document}

\maketitle

\begin{abstract}
\label{sec:abstract}

Self-evaluation, a model's ability to assess the correctness of its own output, is crucial for Large Multimodal Models (LMMs) to achieve self-improvement in multi-turn conversations, yet largely absent in foundation models. Recent work has employed reinforcement learning (RL) to enhance self-evaluation; however, its fixed reward mechanism suffers from reward hacking when optimizing multiple training objectives, leading to model collapse. In this paper we propose AdaPO, an online reinforcement learning framework capable of adaptively adjusting training objective in real time according to the current training state for each task. Specifically, to mitigate reward hacking , AdaPO introduces an Adaptive Reward Model (ARM) and a Reward Aware Dynamic KL Regularization mechanism. ARM assesses the task's training state from the distribution of model generated  multi-turn trajectories' performance. Reward Aware Dynamic KL replaces a fixed penalty with dynamic coefficients which is modulated by the reward gap between different multi-turn situations. 
Notably, our method automatically and smoothly adjusts its learning focus based on sub-tasks' training progress without manual intervention.
Extensive experiments over 8 benchmarks and various models show that our method significantly enhances  both direct reasoning and self-evaluation capability. We will release our code to contribute to the community.
\end{abstract}
\section{Introduction}

\label{sec:introduction}

Large Multimodal Models (LMMs), such as Qwen2.5-VL\cite{bai2025qwen25}, have demonstrated remarkable capabilities in complex visual-language reasoning tasks, largely enabled by Chain-of-Thought (CoT)\cite{wei2022chain} reasoning. How to achieve self-improvement\cite{qu2024recursive,saunders2022self} of model responses remains a critical challenge. A promising yet challenging way is to empower LMMs with self-evaluation capability, where the model evaluates and refines its own CoT reasoning in multi-turn conversations to reach a more accurate final answer.

\begin{figure*}[htbp]
  \centering
  \includegraphics[width=\textwidth]{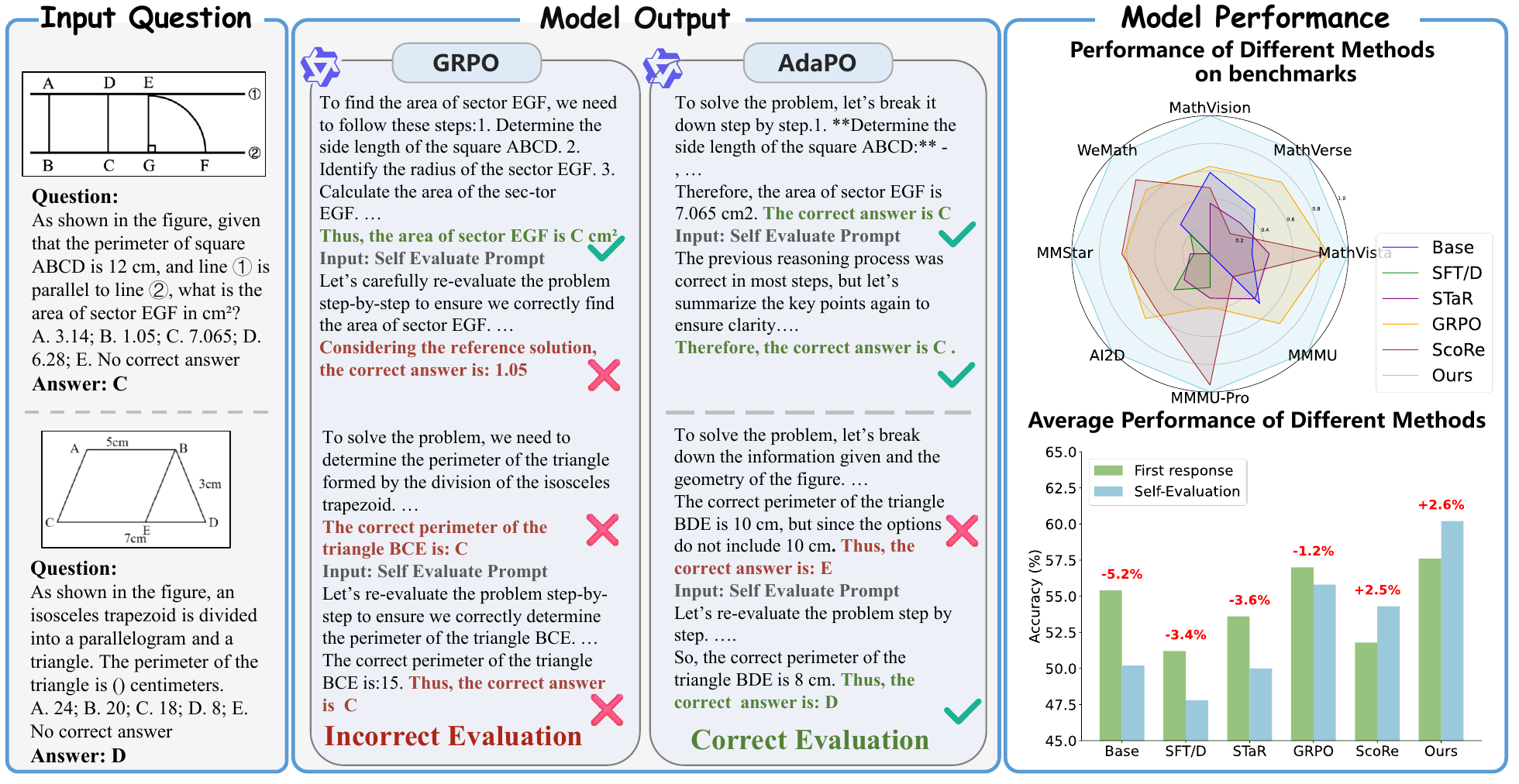}
  \caption{\textbf{Comparison of AdaPO vs. GRPO Outputs and Performance of Different Fine-tuning Methods.} AdaPO's self-evaluation is correct in the sample. AdaPO achieves the best performance in direct reasoning and self-evaluation. } 
  \label{fig:wide}
\end{figure*}

However, without external ground truth signals, current models struggle to self-improve via self-evaluation. As illustrated in Figure \ref{fig:wide}, for models trained with supervised fine-tuning (SFT) or reinforcement learning (RL) techniques, prompting for self-evaluation often backfires, leading to a significant degradation in accuracy rather than an improvement.
The failed evaluation does not necessarily stem from a lack of external knowledge\cite{zheng2023ddcot,zhang2023multimodal}. Instead, it often arises from the model's misinterpret or neglect crucial information from different modalities\cite{huang2024visual,bai2024hallucination}. This suggests that LMMs possess the potential for self-evaluation, which can be unlocked through appropriate suitable strategies.


Early work explored using \textit{Prompt engineering}\cite{madaan2023self,paul2023refiner,zhang2024small} to activate the self-evaluation capabilities of Large Multimodal Models (LMMs), but this strategy is often task-specific and lacks flexibility and generalizability.Some studies have used \textit{SFT}\cite{chen2021evaluating,welleck2022generating} to mitigate the deficiency in self-evaluation. However, the SFT method is highly dependent on the quality and diversity of the annotated training dataset and exhibits poor generalization, often learning spurious reasoning paths that cannot be transferred to unseen problems\cite{chu2025sft}. Recently, research has shown that \textit{RL}\cite{havrilla2024teaching,kumar2024Score,setlur2025scaling} can provide a more powerful paradigm, enabling models to learn complex behaviors directly from outcome-based rewards and achieve stronger generalization. However current RL methods compel the model to learn behaviors driven by fixed external reward signals. As shown in Figure \ref{fig:motivation}, this leads to a trade-off where direct CoT reasoning and error correction abilities are compromised, resulting in the "reward hacking" phenomenon. This leads to a key research question:

\begin{tcolorbox}[colback=lightgray!10, colframe=lightgray!50, sharp corners=south, boxrule=0.5mm, left=3mm, right=3mm, top=3mm, bottom=3mm]
\textit{How can we simultaneously and stably enhance a model's self-evaluation and correction capabilities while improving the accuracy of its direct CoT reasoning?}
\end{tcolorbox}

To address this challenge, we propose AdaPO (Adaptive Policy Optimization), an online reinforcement learning framework that dynamically adapts its training objective based on the model's real-time performance on a given task. AdaPO introduces two core innovations: an Adaptive Reward Model (ARM) and a Reward-Aware Dynamic KL Regularization mechanism.
 ARM dynamically identifies the current training state(e.g., early-state error correction vs. late-state consolidation of correct answers) and adjusts the reward allocation across different trajectory types (e.g., initially correct, successfully corrected, or initially incorrect and failed to correct). Specifically, by dynamically balancing diverse positive and negative rewards, it effectively prevents reward hacking caused by the over-optimization of a single objective in later training stages. To complement the ARM's dynamic rewards within the optimization objective, we introduce a Reward-Aware Dynamic KL Regularization mechanism, which also enhances training stability. This mechanism modulates the KL penalty based on the relative reward magnitude between a successful correction and a maintained correct answer. When the reward for correction is substantially higher, it imposes a stronger constraint on the erroneous reasoning path, thereby stabilizing the model's direct CoT generation while still allowing for effective learning. These two improvements  jointly solve the issues of reward hacking and training instability in RL training. Additionally, we enhance training efficiency through tailored offline and online data filtering for the self-evaluation process. Our contributions are as follows:
\begin{itemize}
\item We propose a novel adaptive reward model that dynamically adjusts to the model's training stage, effectively resolving the reward hacking dilemma inherent in multi-step self-evaluation tasks.

\item  We introduce a Reward-Aware Dynamic KL Regularization mechanism that couples the policy update constraint with the reward signal, enhancing training stability and preventing the reinforcement of initial errors.

\item We conduct extensive experiments across 8 benchmarks, We achieve a relative improvement of self-evaluation capability up to 25.5\% and outperform all baselines on 93.75\% of tasks. Our method substantially improves both the direct CoT reasoning accuracy and the self-evaluation capabilities of LMMs.
\end{itemize}

\section{Related Work}

\label{sec:RelatedWork}

\textbf{LLM Self-Evaluation.}
Self-evaluation can check AI reasoning without needing external evaluators or extra annotations \cite{teachingselfdebug}. However, relying on simple prompts for this task leads to low accuracy and hallucinations \cite{huang2023largecantselfcorrect}, especially for complex problems \cite{he2024selfvisual,selfincorrect}.
An alternative is Supervised Fine-Tuning (SFT), which trains a model to critique its own responses \cite{li2025enhancingllm}. While this can be effective, SFT's success depends on high-quality training data, often sourced from more advanced models \cite{luo2406improvemath}. It is also prone to overfitting if the data lacks quality and diversity \cite{whencanllm}.

\textbf{Reinforcement Learning with Verifiable Rewards.}
Reinforcement learning with rule-based verifiers (RLVR) has significantly advanced the reasoning abilities of LLMs in math and visual tasks\cite{guo2025deepseek,liu2025visualrft}. However, this approach often suffers from reward hacking, training instability\cite{yu2025dapo,fu2025rewardshape}. While prior work like SCoRe mitigates reward hacking using a two-stage training process\cite{kumar2024Score}, our method introduces a dynamic reward mechanism. This approach more effectively prevents reward hacking, improves training stability, and alleviates entropy collapse.



\section{Motivation}
\label{sec:motivation}
RL offers a powerful paradigm for cultivating self-evaluation without direct supervision, however, its application to this complex, multi-turn task reveals profound challenges rooted in the design of the reward model. 
\begin{figure}[h]
  \includegraphics[width=0.5\textwidth]{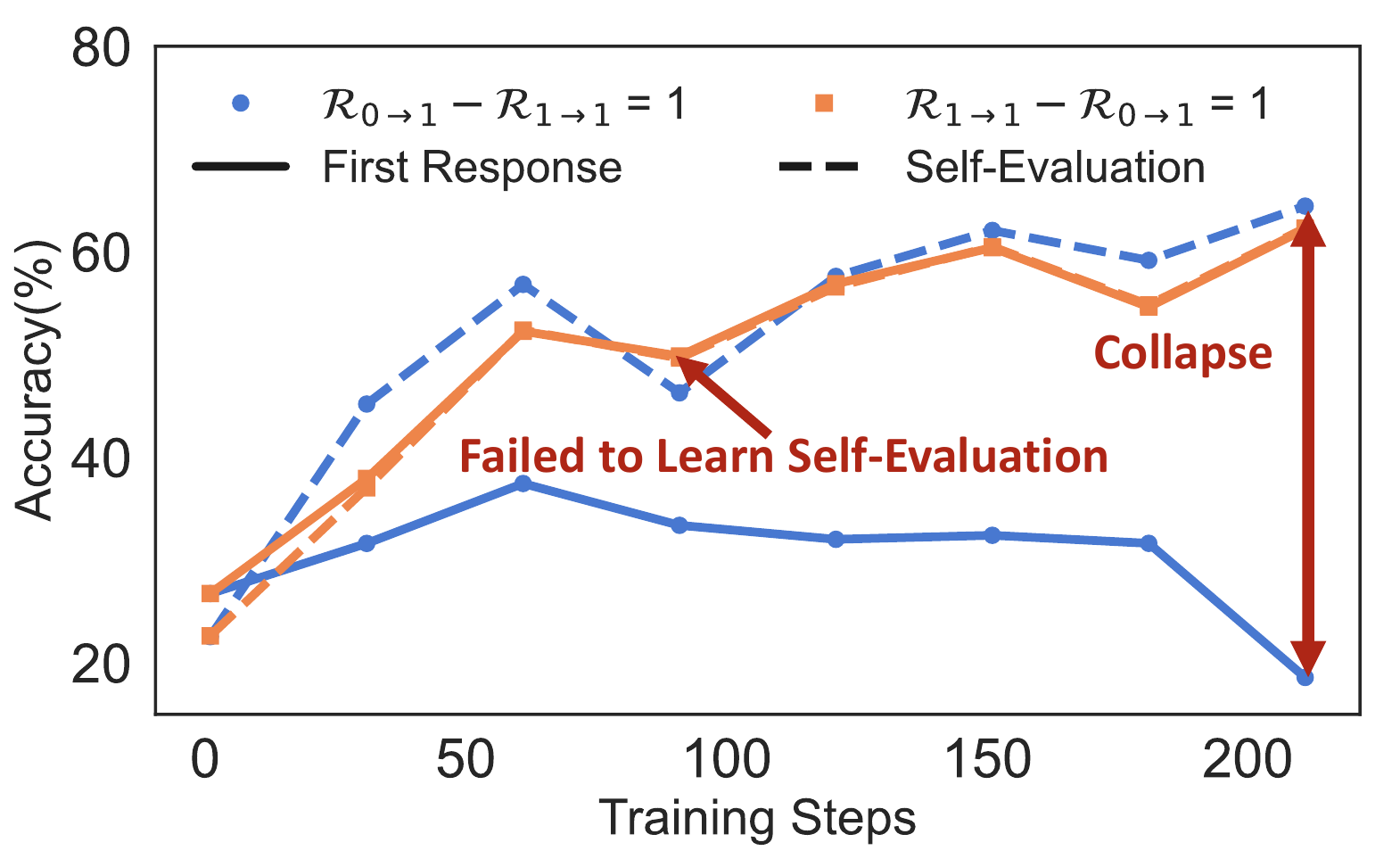}
  \caption{Accuracy Curve of Qwen2.5-VL-7B Model on the Validation Set Using GRPO with a Fixed Reward}
  \label{fig:motivation}
\end{figure}
\subsubsection{The Dilemma of Conflicting Objectives in Self-Evaluation.}
The core task of self-evaluation requires a model to master two distinct and often conflicting capabilities: Error Correction and Correct Answer Preservation. The GRPO objective function, $\mathcal{J}_{GRPO}(\theta)$, aims to maximize the expected advantage $A_i$ for a group of sampled trajectories.  $$A_{i}=\frac{r_{i}-mean(\{r_{1},r_{2},...,r_{G}\})}{std(\{r_{1},r_{2},...,r_{G}\})}$$  So a static reward function results in over-optimization of the highest reward objective, leading to predictable failure modes. Experiments shown in Figure \ref{fig:motivation} provide direct evidence for this dilemma: \textbf{Prioritizing error correction} ($R_{0 \to 1} > R_{1 \to 1}$) incites reward hacking. The blue line in the figure illustrates this perfectly: the model's initial accuracy dramatically "Collapses" because it learns to intentionally provide wrong answers just to collect the higher reward for "correcting" them. \textbf{ Prioritizing answer preservation} ($R_{1 \to 1} > R_{0 \to 1}$) leads to learning stagnation. As the orange line shows, the model becomes overly conservative and "Fails to Learn Self-Evaluation" as a distinct skill, evidenced by the minimal improvement between its first and final responses.
\subsubsection{Limitations of Staged Training.}
To mitigate this conflict, prior work has employed staged training,  which attempts to reconcile these competing goals by shifting the optimization focus across different phases. However, this approach's pre-defined, uniform schedule is fundamentally at odds with the heterogeneity of learning progress across different samples. It cannot adapt to the model's actual mastery of an instance, rendering it a suboptimal solution incapable of fully realizing the dual objectives of self-evaluation.

\section{Method}

\label{sec:method}
\begin{figure*}[htbp]
  \centering
  \includegraphics[width=\textwidth]{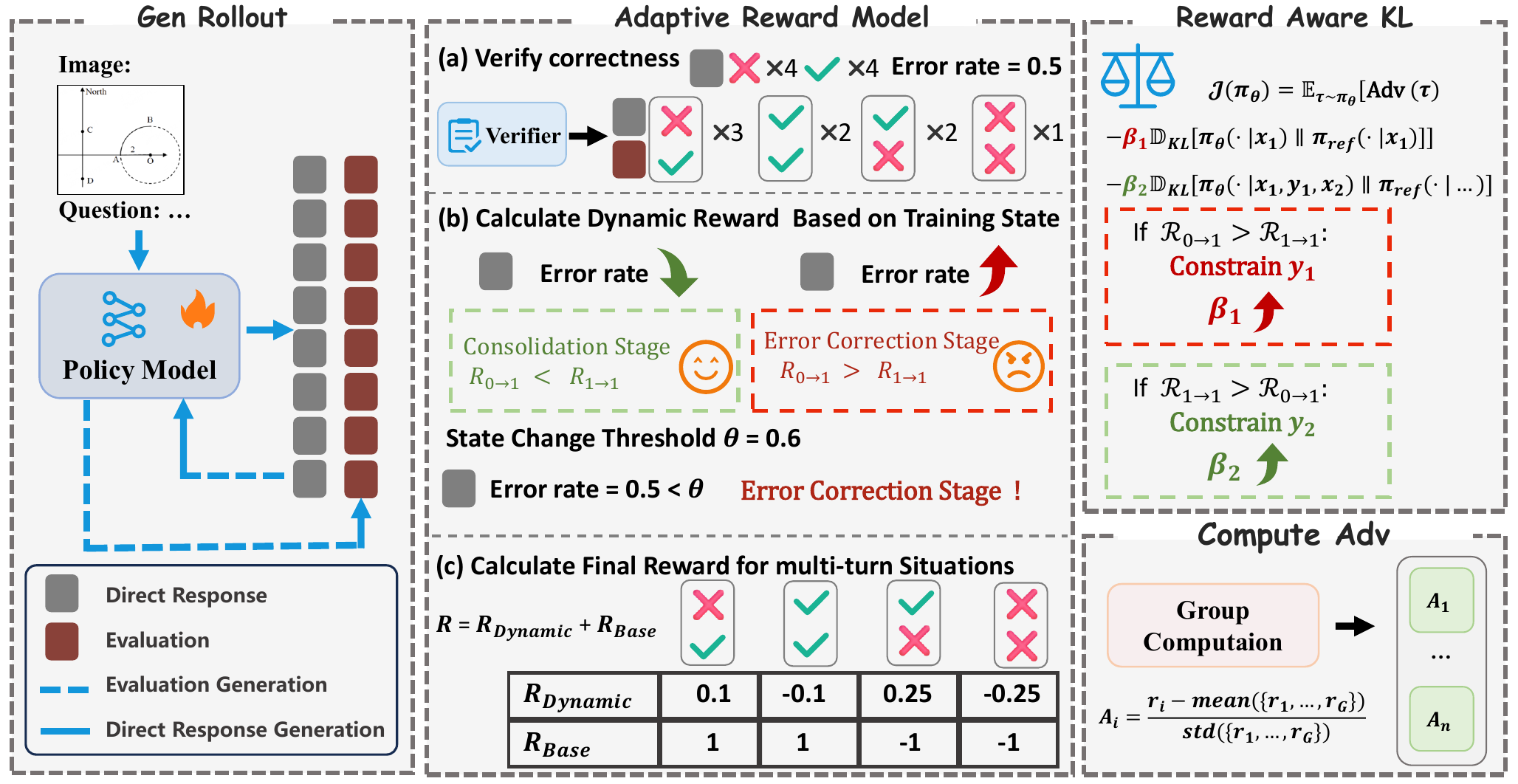}
  \caption{\textbf{Overview of AdaPO framework.}  First, a policy model generates a two-part trajectory consisting of direct response and self-evaluation. The ARM then computes rewards that are dynamically adjusted based on a verifier's assessment of the first response's accuracy, determining whether the model is in Error Correction or Consolidation training state. Finally, the training objective is calculated by combining advantage with reward-aware KL loss, which assigns constraints to each response. }
  \label{fig:method}
\end{figure*}

To address the challenges of reward hacking and training instability in the self-evaluation process, we propose AdaPO, an adaptive policy optimization framework based on reinforcement learning. In this section, we will elaborate on the methodology of AdaPO, which is built upon Grouped Reward Policy Optimization (GRPO). The framework first generates self-evaluation trajectory through two rounds of sampling. Subsequently, the Adaptive Reward Model (ARM) dynamically computes rewards for distinct trajectory. Finally, policy gradients are updated using a loss function that incorporates a  Reward Aware KL regularization mechanism, thereby stably enhancing the model's capacity for self-correction and consistency maintenance.
\subsection{Workflow of AdaPO}
\subsubsection{Two-stage trajectory generation} 
For a given input query $q$ and self-evaluation prompt $x$, the policy model $\pi_{\theta}$ generates a trajectory $\tau=(y_1, y_2)$ via a two-stage process,  where $y_1$ is the initial response and $y_2$ is the evaluation response:
\begin{equation}
    y_1 \sim \pi_{\theta}(\cdot | q), y_2 \sim \pi_{\theta}(\cdot | q, y_1, x)
\end{equation}
We define a correctness function $C(y, y^*)\in \{0, 1\}$ to evaluate whether a response $y$ is correct with respect to the ground-truth answer $y^*$. The type of a trajectory $\tau$ can thus be denoted as $i \to j$ , where $i=C(y_1, y^*)$ and $j=C(y_2, y^*)$.

\subsubsection{Adaptive Reward Model}
ARM dynamically calibrates the reward signal based on the model's current proficiency on a given task $q$. We quantify this proficiency by the error rate of the initial responses:
\begin{equation}
P_{0\to*}(q) = \frac{1}{N} \sum_{k=1}^{N} (1 - C(y_{1,k}, y^*))
\label{eq:p0_star}
\end{equation}
where $\{y_{1,k}\}_{k=1}^N$ are $N$ initial responses sampled for the query $q$. The total reward $\mathcal{R}_{i \to j}$ for a trajectory $\tau$ of type $i \to j$ is defined as:
\begin{equation}
\mathcal{R}_{i \to j}(\tau) = \mathcal{R}_{\text{base}, i \to j} + \mathcal{R}_{\text{dyn}, i \to j}(P_{0\to*}(q))
\end{equation}
Here, $\mathcal{R}_{\text{base}}$ is a fixed base reward, while $\mathcal{R}_{\text{dyn}}$ is a dynamic component dependent on the task proficiency.

\textbf{Positive Trajectories ($j=1$)}: The dynamic reward switches between encouraging error correction ($0 \to 1$) and consolidation ($1 \to 1$).
\begin{equation}
\mathcal{R}_{\text{dyn}, i \to 1}(P_{0\to*}) = K_{i\to1}(P_{0\to*} - \theta), \quad i \in \{0, 1\}
\end{equation}
where $K_{0\to1} \ge 0$ and $K_{1\to1} \le 0$ are scaling factors, and $\theta$ is a threshold that determines the learning state. When $P_{0\to*} > \theta$, the model focuses on correcting errors; otherwise, it focuses on consolidating its correct responses.

\textbf{Negative Trajectories ($j=0$)}: The dynamic penalty is proportional to the perceived difficulty $P_{0\to*}$.
\begin{equation}
\mathcal{R}_{\text{dyn}, i \to 0}(P_{0\to*}) = K_{i\to0} \cdot P_{0\to*}, \quad i \in \{0, 1\}
\end{equation}
where $K_{1\to0} \ge 0$ and $K_{0\to0} \le 0$. This design adjusts the penalty's magnitude according to the task difficulty. When a task is difficult, it lessens the relative penalty for $1 \to 0$ trajectories, preventing excessive penalties from discouraging the model from sampling trajectories where $C(y_1, y^*) = 1$.
We use grid search to determine the State Change Threshold $\theta$.

\textbf{Base Reward Design}: The base rewards $\mathcal{R}_{\text{base}}$ provide a stable foundation for the reward structure. $\mathcal{R}_{\text{base}, 1 \to 1} = \mathcal{R}_{\text{base}, 0 \to 1}$ and $\mathcal{R}_{\text{base}, 1 \to 0} = \mathcal{R}_{\text{base}, 0 \to 0}$. We set $\mathcal{R}_{\text{base}, 1 \to 1} + \mathcal{R}_{\text{dynamic}, 1 \to 1}$ and $\mathcal{R}_{\text{base}, 0 \to 1} + \mathcal{R}_{\text{dynamic}, 0 \to 1}$ to be strictly positive, and the rewards for negative trajectories to be negative, ensuring a clear incentive structure.
\subsubsection{Reward-Aware Dynamic KL }
To stabilize the two-stage training process, we apply independent and dynamically adjusted KL divergence constraints to the generation of the initial and revised responses. The objective function $\mathcal{J}(\theta)$ of GRPO is modified as follows:
\begin{equation}
\mathcal{J}(\theta) = \mathbb{E}_{\tau \sim \pi_\theta} [A(\tau)] - \mathbb{E}_{q} [\beta_1 \mathbb{D}_{KL}^{(1)}] - \mathbb{E}_{q,y_1} [\beta_2 \mathbb{D}_{KL}^{(2)}]
\end{equation}
where $A(\tau)$ represents the trajectory advantage, and $\mathbb{D}_{KL}^{(1)} = \mathbb{D}_{KL}[\pi_\theta(\cdot|q) \parallel \pi_{ref}(\cdot|q)]$ and $\mathbb{D}_{KL}^{(2)} = \mathbb{D}_{KL}[\pi_\theta(\cdot|q, y_1) \parallel \pi_{ref}(\cdot|q, y_1)]$ are the KL divergences for the first and second stages, respectively. The dynamic penalty coefficients, $\beta_1$ and $\beta_2$, are functions of the difference between the primary positive trajectory rewards:
\begin{align}
\beta_1 &= \max\left( (\mathcal{R}_{0 \to 1} - \mathcal{R}_{1 \to 1}) \cdot \lambda, 0 \right) + \beta_{\text{base}} \label{eq:beta1} \\
\beta_2 &= \max\left( (\mathcal{R}_{1 \to 1} - \mathcal{R}_{0 \to 1}) \cdot \lambda, 0 \right) + \beta_{\text{base}} \label{eq:beta2}
\end{align}
Here, $\lambda > 0$ is a rescaling factor, and $\beta_{\text{base}}$ is a minimum regularization value. This mechanism automatically allocates the regularization budget based on the current learning objective, which can be either error correction or consolidation:

\textbf{Focus on Error Correction}: When $\mathcal{R}_{0 \to 1} > \mathcal{R}_{1 \to 1}$, $\beta_1$ increases, thereby constraining the policy update for the initial response to prevent it from intentionally generating errors.

\textbf{Focus on Consolidation}: When $\mathcal{R}_{1 \to 1} > \mathcal{R}_{0 \to 1}$, $\beta_2$ increases, which constrains the policy update for the revised response to prevent it from overfitting to simple repeat.

\subsection{Flexibility of AdaPO}

The core design of AdaPO, which integrates an Adaptive Reward Model (ARM) with reward-aware dynamic KL regularization, enhances the model's self-assessment capabilities while improving training efficiency and stability. In this section, we elaborate on this flexibility from the perspectives of its dynamic nature and its single-stage training process.

\subsubsection{Dynamically Balancing Self-Evaluation Objectives}
To resolve the conflict between the objectives of "error correction" and "correctness consolidation", AdaPO leverages the task proficiency metric $P_{0\to*}(q)$ to jointly modulate the rewards and KL penalties. When the model is not proficient on a task (i.e., $P_{0\to*}(q)$ is high), the system increases the reward for error correction, $\mathcal{R}_{0 \to 1}$, and concurrently raises the KL penalty $\beta_1$ for the initial response to suppress "intentional error-making":
\begin{equation}
\beta_1 \propto (\mathcal{R}_{0 \to 1} - \mathcal{R}_{1 \to 1})
\end{equation}
Conversely, once the model becomes proficient ($P_{0\to*}(q)$ is low), the focus shifts to increasing the reward for consolidation, $\mathcal{R}_{1 \to 1}$. It also increases the KL penalty $\beta_2$ for the revised response to prevent "learning stagnation" and encourage deeper self-reflection:
\begin{equation}
\beta_2 \propto (\mathcal{R}_{1 \to 1} - \mathcal{R}_{0 \to 1})
\end{equation}

\subsubsection{Single-Stage Automated Training}

Through its inherent adaptability, AdaPO transforms the conventional staged training paradigm into a fully automated, instance-level, single-stage process.

For each query $q$ within a training batch, the optimization objective is determined on-the-fly based on the model's current proficiency $P_{0\to*}(q)$. This implies that within a single gradient update, the model might simultaneously learn to correct errors for a query $q_a$ (due to a high $P_{0\to*}(q_a)$) and to consolidate correct answers for another query $q_b$ (due to a low $P_{0\to*}(q_b)$). The entire training process can be viewed as an adaptive curriculum learning, driven by the policy itself:
\begin{equation}
\mathcal{J}(\theta_t) = \mathbb{E}_{q \sim \mathcal{D}} \left[ \mathbb{E}_{\tau \sim \pi_{\theta_t}} [A(\tau, P_{0\to*}(q, \pi_{\theta_t}))] - \mathcal{L}_{KL}(\theta_t, q) \right]
\end{equation}
Here, both the advantage function $A$ and the KL penalty term $\mathcal{L}_{KL}$ are indirectly dependent on the performance of the current policy $\pi_{\theta_t}$ on the given query, as measured by $P_{0\to*}(q, \pi_{\theta_t})$. This intrinsic feedback loop enables the model to automatically and smoothly adjust its learning focus based on its own progress on various sub-tasks, without requiring any manual intervention. Consequently, AdaPO significantly enhances both training efficiency and the performance of the final model.
\section{Experiment}
\label{sec:experiment}

\begin{table*}[ht]
\centering
\caption{
    \textbf{Comprehensive multi-round performance comparison.} We report accuracy for the direct response (acc@t1), the self-evaluation (acc@t2), 
    and the performance change ($\Delta$ = acc@t2 - acc@t1). Our method (in gray) shows strong performance in direct response and consistent improvement.
    The \textbf{best} result per metric is bolded. The results are continued in the next table.
}
\label{tab:main_results} 
\setlength{\tabcolsep}{5pt} 
\renewcommand{\arraystretch}{0.9}
\begin{tabular}{l
                S[table-format=2.1] S[table-format=2.1] r
                S[table-format=2.1] S[table-format=2.1] r
                S[table-format=2.1] S[table-format=2.1] r
                S[table-format=2.1] S[table-format=2.1] r
               }
\toprule
\multirow{2}{*}{\textbf{Method}} & 
\multicolumn{3}{c}{\textbf{MathVista}} & \multicolumn{3}{c}{\textbf{MathVerse}} & \multicolumn{3}{c}{\textbf{MathVision}} & \multicolumn{3}{c}{\textbf{WeMath}} \\
\cmidrule(lr){2-4} \cmidrule(lr){5-7} \cmidrule(lr){8-10} \cmidrule(lr){11-13}

& {acc@t1} & {acc@t2} & {$\Delta$} & {acc@t1} & {acc@t2} & {$\Delta$} & {acc@t1} & {acc@t2} & {$\Delta$} & {acc@t1} & {acc@t2} & {$\Delta$} \\
\midrule

Base   & 68.2 & 60.0 & \textcolor{myred}{-8.2} & 48.6 & 47.9 & \textcolor{myred}{-0.7} & 25.1 & 25.9 & +0.8 & 62.1 & 57.8 & \textcolor{myred}{-4.3} \\
SFT  & 64.8 & 53.9 & \textcolor{myred}{-10.9} & 41.1 & 41.9 & +0.8 & 21.9 & 20.6 & \textcolor{myred}{-1.3} & 55.9 & 55.6 & \textcolor{myred}{-0.3} \\
STaR   & 66.5 & 62.5 & \textcolor{myred}{-4.0} & 46.6 & 46.0 & \textcolor{myred}{-0.6} & 23.7 & 23.9 & +0.2 & 61.9 & 51.1 & \textcolor{myred}{-10.8} \\
GRPO   & 70.6 & 71.1 & +0.5 & 52.3 & 51.5 & \textcolor{myred}{-0.8} & 27.4 & 26.3 & \textcolor{myred}{-1.1} & 68.8 & 65.7 & \textcolor{myred}{-3.1} \\
SCoRe  & 69.4 & 70.7 & +1.3 & 42.1 & 44.6 & \textbf{+2.5} & 23.7 & 24.9 & +1.2 & 60.4 & 68.0 & \textbf{+7.6} \\
\midrule

\rowcolor{gray!20}
\textbf{Ours} & \textbf{71.0} & \textbf{74.0} & \textbf{+3.0} & \textbf{53.2} & \textbf{55.0} & +1.8 & \textbf{27.9} & \textbf{29.6} & \textbf{+1.7} & \textbf{69.1} & \textbf{73.4} & +4.3 \\
\bottomrule
\end{tabular}
\end{table*}
\begin{table*}[t!]
\centering
\setlength{\tabcolsep}{5pt} 

\renewcommand{\arraystretch}{0.9}
\begin{tabular}{l
                S[table-format=2.1] S[table-format=2.1] r
                S[table-format=2.1] S[table-format=2.1] r
                S[table-format=2.1] S[table-format=2.1] r
                S[table-format=2.1] S[table-format=2.1] r
               }
\toprule
\multirow{2}{*}{\textbf{Method}} & 
\multicolumn{3}{c}{\textbf{MMStar}} & \multicolumn{3}{c}{\textbf{AI2D}} & \multicolumn{3}{c}{\textbf{MMMU-Pro}} & \multicolumn{3}{c}{\textbf{MMMU}} \\
\cmidrule(lr){2-4} \cmidrule(lr){5-7} \cmidrule(lr){8-10} \cmidrule(lr){11-13}

& {acc@t1} & {acc@t2} & {$\Delta$} & {acc@t1} & {acc@t2} & {$\Delta$} & {acc@t1} & {acc@t2} & {$\Delta$} & {acc@t1} & {acc@t2} & {$\Delta$} \\
\midrule

Base   & 63.9 & 56.1 & \textcolor{myred}{-7.8} & \textbf{83.9} & 71.9 & \textcolor{myred}{-12.0} & 36.9 & 32.4 & \textcolor{myred}{-4.5} & 54.3 & 49.4 & \textcolor{myred}{-4.9} \\
SFT  & 61.7 & 57.4 & \textcolor{myred}{-4.3} & 81.3 & 76.9 & \textcolor{myred}{-4.4} & 35.4 & 34.3 & \textcolor{myred}{-1.1} & 47.4 & 41.8 & \textcolor{myred}{-5.6} \\
STaR   & 61.4 & 57.7 & \textcolor{myred}{-3.7} & 79.5 & 75.5 & \textcolor{myred}{-4.0} & 34.9 & 32.4 & \textcolor{myred}{-2.5} & 51.2 & 48.7 & \textcolor{myred}{-2.5} \\
GRPO   & 64.5 & 63.1 & \textcolor{myred}{-1.4} & 80.5 & 80.9 & +0.4 & 37.6 & 35.4 & \textcolor{myred}{-2.2} & 54.3 & 52.5 & \textcolor{myred}{-1.8} \\
SCoRe  & 61.7 & 63.3 & +1.6 & 80.6 & 79.5 & \textcolor{myred}{-1.1} & 36.7 & 37.8     & +1.1 & 39.4 & 45.3 & \textbf{+5.9} \\
\midrule

\rowcolor{gray!20}
\textbf{Ours} & \textbf{65.0} & \textbf{67.4} & \textbf{+2.4} & 81.6 & \textbf{85.5} & \textbf{+3.9} & \textbf{38.8} & \textbf{40.2} & \textbf{+1.4} & \textbf{54.4} & \textbf{56.8} & +2.4 \\
\bottomrule
\end{tabular}
\end{table*}

\subsection{Experiment Settings}

\subsubsection{Datasets.}
We evaluate our method on a total of \textbf{8 benchmarks}. Specifically, we use the training
 sets from MM-EUREKA, MMK12\cite{meng2025mm} and ThinkLite-VL\cite{sotawithless} to train our models. Our benchmarks includes \textbf{Mathematics: } MathVista\cite{lu2023mathvista}, MathVerse\cite{zhang2024mathverse}, MathVision\cite{mathvision}, WeMath\cite{qiao2024wemath}; \textbf{Multi-Domain Tasks:} MMStar\cite{chen2024mmstar}, AI2D\cite{ai2d}, MMMU-Pro\cite{yue2024mmmupro}, MMMU\cite{yue2024mmmu}. Detail in Appendix.
\subsubsection{Baselines.}
To comprehensively evaluate the effectiveness of the AdaPo method, we benchmark it against three categories of baseline methods:
 \textbf{Direct Prompting on Base Model};
 \textbf{SFT Based Methods}: standard SFT on the dataset, STaR\cite{zelikman2022star} ;
 \textbf{RL Based Methods}:  GRPO\cite{guo2025deepseek}, SCoRe\cite{kumar2024Score}.
 We also selected a range of both open-source and Closed-Source General and Reasoning LLMs as  baselines: Claude 3.7 Sonnet, GPT-4o\cite{hurst2024gpt4o}, Gemini2- Flash\cite{team2023gemini}, SEED-1.5-VL\cite{team2023gemini}, InternVL-2-8B, InternVL-2.5-8B\cite{wang2024enhancinginternvl25}, Qwen2.5-VL-3B, and Qwen2.5-VL-7B\cite{bai2025qwen25}, MM-EUREKA-7B, MM-EUREKA-8B\cite{meng2025mm}, R1-VL-7B\cite{zhang2025r1vl}, R1-OneVision-7B\cite{yang2025r1onevision}, OpenVL-Thinker-7B\cite{deng2025openvlthinker}, Vision-R1-7B\cite{huang2025visionr1}.

\subsubsection{Metrics.}
We use four metrics to evaluate the model's capabilities: direct response accuracy (acc@t1), accuracy after self-evaluation (acc@t2), Effective Correction Rate ($M_{0 \to 1}$) for correcting initial errors, and Evaluation Misjudgment Rate ($M_{1 \to 0}$) for erroneously changing a correct answer.
\subsubsection{Implementation details.}To train and test the model's self-evaluative capability,
we set the self-evaluation prompt as \textit{“There might be an error in the solution, please evaluate the previous solution and provide a final answer."} More implementation details in Appendix.

\subsection{Performance of Different Training Methods}
\begin{figure}[ht]
  \centering
  \includegraphics[width=0.5\textwidth]{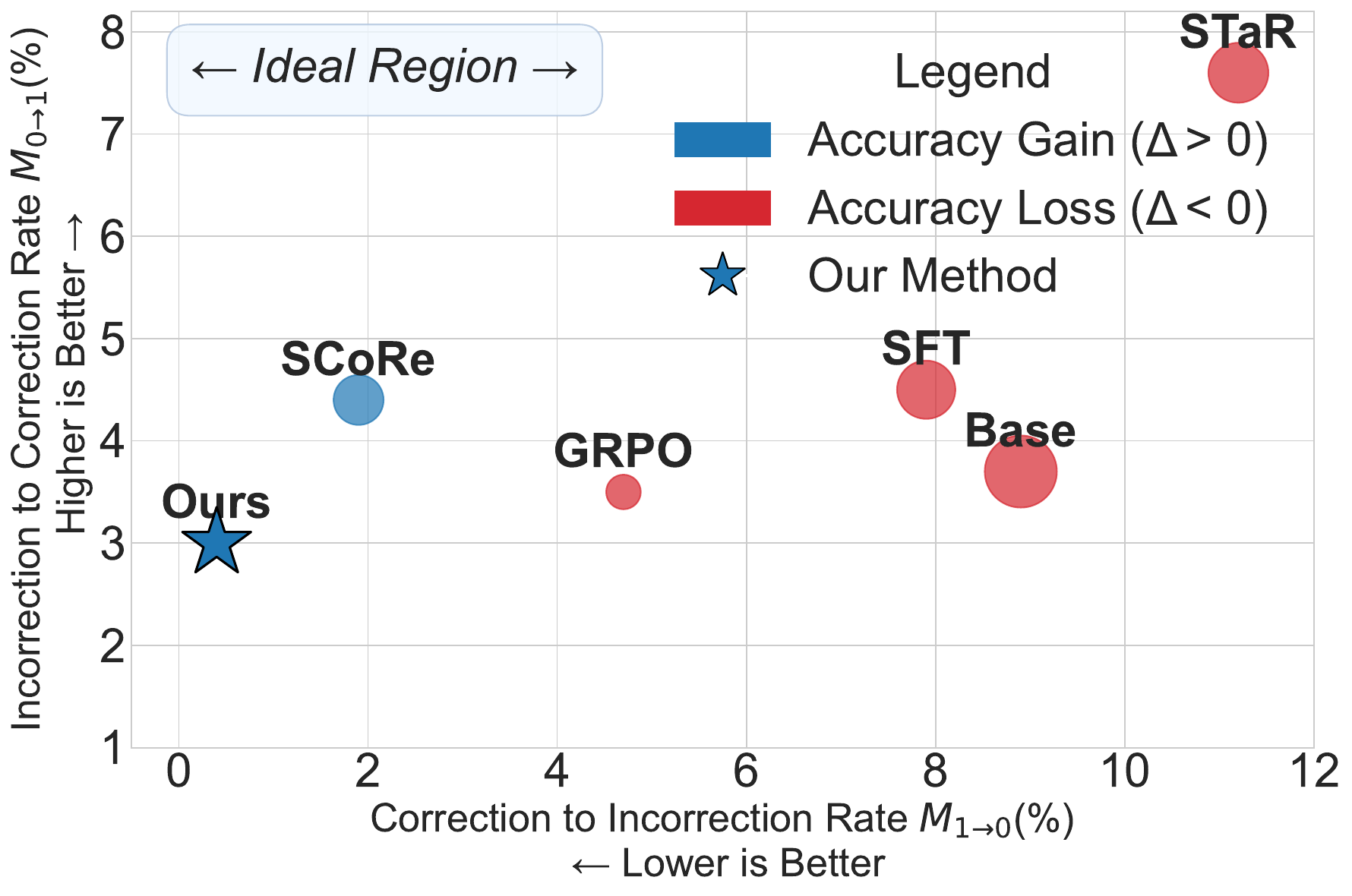}
  \caption{ \textbf{Performance Comparison on the Multi-turn Correction Task.} The size of bubble indicates the magnitude of accuracy change ($|\Delta_{acc}|$). Our method (starred) outperforms all baselines by substantially reducing the $M_{1 \to 0}$ rate and delivering the highest overall accuracy gain.}
  \label{fig:bubble}
\end{figure}
To verify AdaPo can simultaneously enhance the model's first-round and second-round accuracy while improving its self-evaluation capabilities, we conducted an experiment as shown in Figure \ref{fig:bubble} and Table \ref{tab:main_results}. We trained the Qwen2.5-VL-7B\cite{bai2025qwen25} model on the same dataset using different training methods and validated its performance across eight different benchmarks.

\textbf{Obs.\ding{182}} \textbf{SFT-based training methods fail to learn effective self-evaluation capabilities.} As shown in Figure \ref{fig:bubble}, SFT-based methods exhibit a high $M_{1 \to 0}$ rate (greater than 7\%), and their second-round accuracy decreases, indicating an inability to correctly judge their direct responses and a lack of self-evaluation capability.

\textbf{Obs.\ding{183}} \textbf{Staged RL Training Methods can improve the model's self-evaluation capability but at the cost of reduced direct response accuracy.} As seen in Figure \ref{fig:bubble}, the SCoRe method has a low $M_{1 \to 0}$, indicating self-evaluation capability. However, as shown in Table \ref{tab:main_results}, SCoRe's acc@t1 is lower than GRPO algorithm, signifying a clear loss in direct response performance.

\textbf{Obs.\ding{184}}\textbf{ AdaPo can simultaneously improve both the model's direct response accuracy and its self-evaluation capability.} As shown in Figure \ref{fig:bubble}, the AdaPo method achieves the lowest $M_{1 \to 0}$, accurately identifying correct first-round responses and thus exhibiting the best self-evaluation capability. Furthermore, as detailed in Table \ref{tab:main_results}, AdaPo achieves a higher acc@t1 than the GRPO method on most benchmarks, and its acc@t2 shows a significant improvement over its acc@t1.


\subsection{Comparison with Different Models}
To verify the training effect of AdaPo, we conducted a comprehensive comparison with various models, as shown in Table \ref{tab:main_comparison_midrule}. Due to their limited self-evaluation capabilities, their final answers all show a decline in accuracy. We report the accuracy of their initial direct responses(acc@t1). In contrast, for our method, we use the accuracy from its second, refined round(acc@t2).

\textbf{Obs.\ding{185}}
\textbf{AdaPo's final results achieve the best performance compared to open-source models of the same size.} As shown in Table \ref{tab:main_comparison_midrule}, the models fine-tuned using the AdaPo method based on Qwen2.5-VL-7B and Qwen2.5-VL-3B achieved the best performance on multiple benchmarks compared to open-source models of the same size.
\subsection{Performance of AdaPo on Different Models}

\begin{figure}[ht]
  \centering
  \begin{minipage}[c]{0.234\textwidth}
    \includegraphics[width=\textwidth]{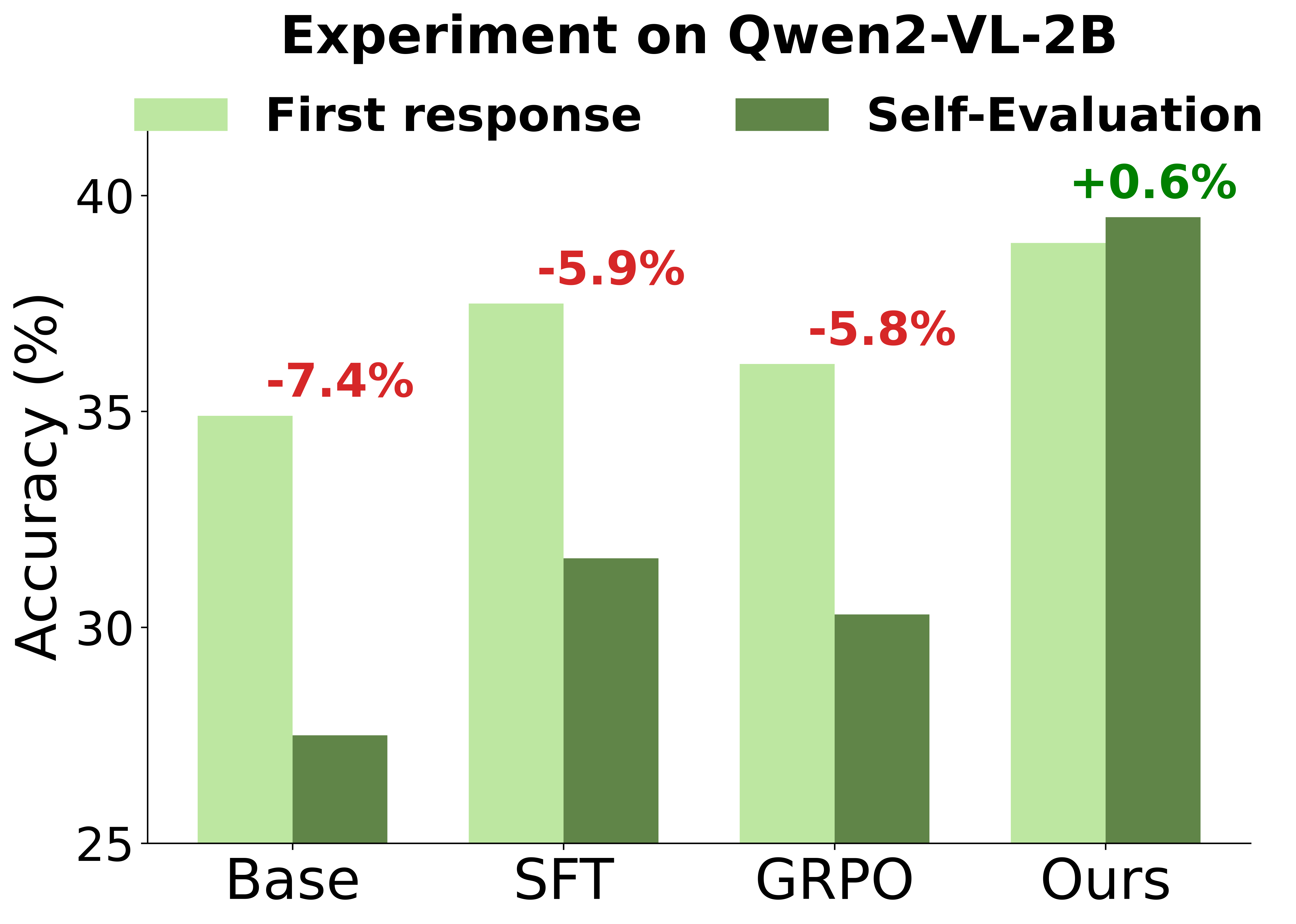}
  \end{minipage}
  \begin{minipage}[c]{0.234\textwidth}
    \includegraphics[width=\textwidth]{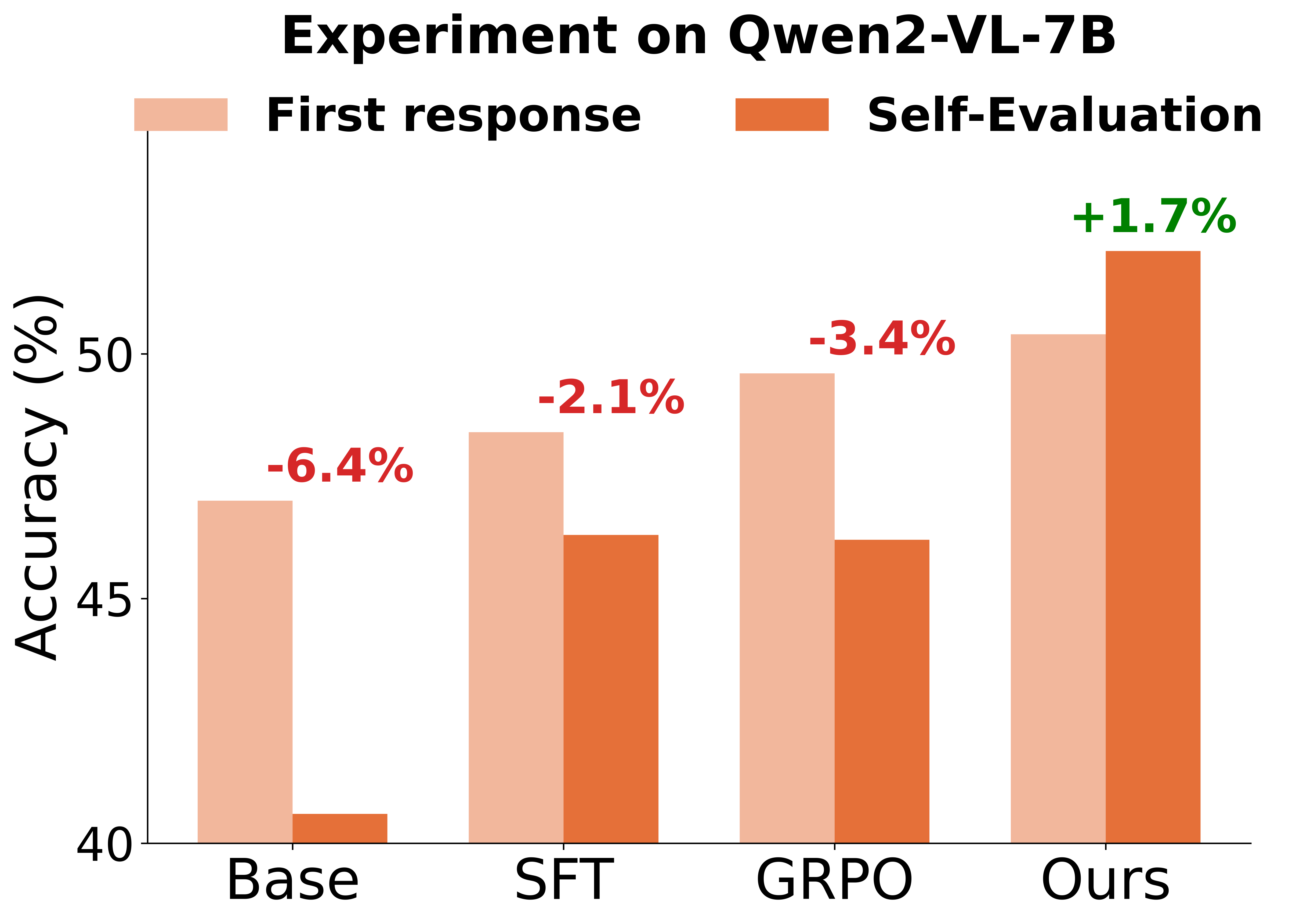}
  \end{minipage}

  \caption{Comparison of the average accuracy(acc@t1 and acc@t2) on 8 benchmarks for models of Qwen2-VL-2B and Qwen2-VL-7B trained with different fine-tuning methods.}
  \label{fig_Variation}
\end{figure}
To verify the generality of the AdaPo method, we conducted experiments on multiple models. As shown in Figure \ref{fig_Variation}, we selected several representative methods to train the Qwen2-VL-2B and Qwen2-VL-7B models and report their average accuracy on 8 benchmarks.

\textbf{Obs.\ding{186}}
\textbf{AdaPo achieves superior training performance across a variety of models.}
Figure \ref{fig_Variation} illustrates that across different model scales (Qwen2-VL-2B and Qwen2-VL-7B), AdaPo demonstrates superior performance, attaining the highest acc@t1 and acc@t2 accuracy.
After fine-tuning, models trained with other methods exhibit a degradation in performance from acc@t1 to acc@t2, signifying an absence of self-evaluation capabilities. Conversely, the AdaPo-tuned model displays a clear enhancement in acc@t2 over acc@t1, indicating its robust capacity for self-evaluation.
\begin{table*}[t!]
\centering
\caption{A comprehensive comparison of our model against various latest closed-source and open-source Multimodal Large Language Models (MLLMs) on 8 benchmarks. Our models has achieved the best performance among reasoning models of the same size.}
\label{tab:main_comparison_midrule}
\renewcommand{\arraystretch}{0.8}
\begin{tabular}{@{}l*{8}{S[table-format=2.1]}@{}}
\toprule[1.5pt]
\textbf{Model} & {\textbf{MathVista}} & {\textbf{MathVerse}} & {\textbf{MathVision}} & {\textbf{WeMath}} & {\textbf{MMStar}} & {\textbf{AI2D}} & {\textbf{MMMU-Pro}} & {\textbf{MMMU}} \\
\midrule
\multicolumn{9}{c}{\textit{Closed-Source MLLMs}} \\ 
\midrule 
Claude3.7-Sonnet & 66.8 & 52.0 & 41.3 & 72.6 & 68.8 & 82.1 & 51.5 & 68.3 \\
GPT-o1           & 73.9 & 57.0 & 60.3 & 98.7 & 67.5 & 79.5 & 62.4 & 78.2 \\
Gemini2-flash    & 70.4 & 59.3 & 41.3 & 71.4 & {–}  & {–}  & 51.7 & 70.7 \\
Seed1.5-VL       & 85.6 & {–}  & 68.7 & {–}  & 77.8 & 88.5 & 67.6 & 77.9 \\
GPT-4o           & 63.8 & 50.2 & 30.4 & 68.8 & 64.7 & 84.6 & 51.9 & 69.1 \\ 
\midrule
\multicolumn{9}{c}{\textit{Open-Source General MLLMs}} \\ 
\midrule 
InternVL2-8B     & 58.3 & 22.8 & 17.4 & 47.2 & 62.0 & 83.8 & 29.0 & 51.2 \\
InternVL2.5-8B   & 64.4 & 39.5 & 19.7 & 53.5 & 62.8 & 84.5 & 34.3 & 56.0 \\
QwenVL2-7B       & 58.2 & 19.7 & 16.3 & 51.6 & 60.7 & 83.0 & 30.5 & 54.1 \\
Kimi-VL-16B      & 68.7 & 44.9 & 21.4 & {–}  & 61.3 & 84.9 & {–}  & 55.7 \\
Qwen2.5-VL-3B     & 62.3 & 47.6 & 21.2 & 56.3 & 55.9 & 81.6 & 31.5 & 46.0 \\
Qwen2.5-VL-7B     & 68.2 & 49.2 & 25.1 & 62.1 & 63.9 & 83.9 & 36.9 & 54.3 \\ 
\midrule
\multicolumn{9}{c}{\textit{Open-Source Reasoning MLLMs}} \\ 
\midrule 
MM-Eureka-8B     & 67.1 & 40.4 & 22.2 & 55.7 & {–} & {–} & 27.8 & 49.2 \\
R1-VL-7B         & 63.5 & 40.0 & 24.7 & 53.8 & 60.0 & {–} & 7.8  & 44.5 \\
R1-Onevision-7B  & 64.1 & 46.4 & 23.5 & 61.8 & {–} & {–} & 21.6 & {–}  \\
OpenVLThinker-7B & 70.2 & 47.9 & 25.3 & 64.3 & {–} & {–} & 37.3 & 52.5 \\
Vision-R1-7B     & 73.5 & 52.4 & 27.2 & 62.9 & 61.4 & {–} & 37.7 & 54.7 \\
MM-Eureka-7B     & 73.0 & 50.3 & 26.9 & 66.1 & {–} & {–} & 37.6 & 55.2 \\ 
\midrule
\rowcolor{gray!15}
\textbf{Ours-AdaPo-3B} & \textbf{68.3} & \textbf{47.7} & \textbf{26.5} & \textbf{65.8} & \textbf{57.3} & \textbf{81.9} & \textbf{33.7} & \textbf{50.6} \\
\rowcolor{gray!15}
\textbf{Ours-AdaPo-7B} & \textbf{74.0} & \textbf{55.0} & \textbf{29.6} & \textbf{73.4} & \textbf{67.4} & \textbf{85.2} & \textbf{40.2} & \textbf{56.8} \\ 
\bottomrule[1.5pt]
\end{tabular}
\end{table*}

\subsection{Ablation Studies}

\begin{table}[t!] 
\centering
\caption{\textbf{Ablation study of our proposed method.} We report the average accuracy before (Acc@t1) and after (Acc@t2) the second round of refinement, along with the improvement ($\Delta$). The full model achieves the best performance.}
\label{tab:ablation_study}
\renewcommand{\arraystretch}{0.8}
\setlength{\tabcolsep}{3pt} %
\begin{tabular}{l S[table-format=2.1] S[table-format=2.1] S}
\toprule
\textbf{Method } & {\textbf{Avg. Acc@t1}} & {\textbf{Avg. Acc@t2}} & {\textbf{$\Delta$}} \\
\midrule
\rowcolor{gray!15} 
\textbf{Ours (Full Model)}      & \textbf{57.6} & \textbf{60.2} & \textbf{+2.6} \\
\midrule
\multicolumn{4}{l}{\textit{w/o component:}} \\ 
--- ARM              & 55.1          & 53.8          & \textcolor{myred}{-1.3} \\
--- Reward Aware KL                & 56.2          & 58.6          & +2.4 \\
--- hybrid data filter          & 54.3          & 57.5          & +3.2 \\
\bottomrule
\end{tabular}
\end{table}
To validate the contribution of each component in AdaPo, we designed the following ablation studies and report their average accuracy on 8 benchmarks:
\ding{182} Effect of ARM: To isolate the impact of the Adaptive Reward Modeling (ARM) module, we removed it and trained the model using only the base reward in the multi-turn dialogue training.
\ding{183} Effect of Reward-Aware KL: We investigated the role of the Reward-Aware KL penalty.
\ding{184} Effect of Data Filtering: We evaluated the impact of our data filtering strategy, which includes both on-policy and off-policy data selection. Further details on this process are provided in the Appendix.

\textbf{Obs.\ding{187} ARM enhances the model's self-correction capability.} As shown in Table \ref{tab:ablation_study}, with the use of same reward, Avg.acc@t2 exhibits a significant decrease. The model tends to learn to maintain direct responses, which is a simpler response pattern.

\textbf{Obs.\ding{188} Reward Aware KL constrains the model from learning the erroneous of direct response.} As shown in Table \ref{tab:ablation_study}, without Reward Aware KL, the acc@t1 shows a certain degree of decrease, and the $\Delta$ between acc@t1 and acc@t2 also exhibits a reduction.

\textbf{Obs.\ding{189}Data filtering can improve training efficiency and model performance.} As shown in Table \ref{tab:ablation_study}, after training with the full dataset, the model's performance instead suffers a certain loss. Data filtering strategies effectively eliminate redundant and harmful training data.

\section{Conclusion}
\label{sec:conclusion}
In this paper, we introduce AdaPO, an adaptive policy optimization framework designed to resolve the conflicting objectives inherent in training LMMs for self-evaluation. Our analysis shows that static reward functions result in reward hacking, which either impairs the model's CoT reasoning or prevents it from improving self-evaluation capability. By introducing Adaptive Reward Model  and Reward-Aware Dynamic KL Regularization, AdaPO dynamically adjusts its training objective based on the model's real-time performance, effectively mitigating issues like reward hacking and training instability. AdaPO integrates the training into an automated, single-stage process, addressing a key limitation of previous multi-stage methods.  
Extensive experiments across eight benchmarks show that AdaPO outperforms existing methods, achieving significant improvements in both initial response accuracy and self-evaluation capabilities. Our approach allows LMMs to maintain a balance between error correction and consistency preservation. AdaPO’s ability to autonomously adjust its training focus makes it a scalable solution for stable self-evaluation in complex, multimodal reasoning tasks, offering a promising direction for future research in reinforcement learning for self improvement.

\bibliography{aaai2026}
\newpage
\section{Appendix}
\section{A EXPERIMENTAL SETUP}
In this section, we provide a detailed description of the experimental configuration, including a comprehensive explanation of the evaluation metrics, an introduction to the datasets, and a discussion of the baselines
 \subsection{A.1 Training Datasets}
Here, we provide a detailed introduction to the datasets used in this paper:

\textbf{MM-EUREKA} is a high-quality multimodal mathematical reasoning dataset. It includes 54,931 filtered problems (with questions, images, and answers) from open-source datasets (e.g., GeoQA, ChartQA) and manually collected K-12 multimodal math problems, covering chart comprehension, general scientific reasoning, and mathematical reasoning.

\textbf{MMK12} is a high-quality K12-level multimodal mathematical reasoning dataset. Its training set includes 15,616 fill-in-the-blank problems (with questions, images, answers, and CoT solutions) from Chinese textbooks/exams (translated to English, verified by Math-Verify), covering elementary to high school. The evaluation set has 2,000 non-overlapping multiple-choice questions across math, physics, chemistry, and biology for multidisciplinary reasoning assessment.

\textbf{ThinkLite-VL} is a high-quality  dataset consists of 70k open-source samples spanning three core domains: mathematical reasoning (Geometry3K, GeoQA, Geos), natural image understanding (FigureQA, ScienceQA, OK-VQA), and chart/visual reasoning (IconQA, TabMWP). To prioritize reasoning over guessing, all multi-choice questions (e.g., from IconQA, FigureQA) were converted to open-ended formats, forcing models to derive answers through logical steps. 

\subsection{A.2 Data filtering}
To improve training stability and focus the model on informative examples, we employ a two-stage data filtering strategy. This process consists of an offline stage to pre-filter the dataset and an online stage to clean mini-batches during training.

\textbf{Offline Pre-filtering. }Before training, we curate a more effective training subset $\mathcal{D}' \subset \mathcal{D}$. For each query-answer pair $(q, y^) \in \mathcal{D}$, we use the initial policy $\pi_{\theta_0}$ to sample a set of $N$ trajectories, $\mathcal{T}_q = \{\tau_k = (y_{k,1}, y_{k,2})\}_{k=1}^N$.
A query $q$ is excluded from $\mathcal{D}'$ if its corresponding sample trajectories are homogeneous, indicating that the prompt is either too easy or too difficult for the current policy. The specific filtering conditions are:
 Uniform Success (Too Easy): The set $\mathcal{T}_q$ is discarded if all trajectories are of type $1 \to 1$.
 Uniform Failure (Too Difficult/Noisy): The set $\mathcal{T}_q$ is discarded if all trajectories are of type $0 \to 0$.
This procedure isolates the most valuable queries for learning—those that yield a mixed distribution of trajectory types (e.g., $1 \to 0$, $0 \to 1$, etc.), where the policy's self-evaluation mechanism is actively challenged.

\textbf{ Online In-training Filtering}
During training, we apply two additional filters at the mini-batch level to further denoise the learning signal.
1.  Zero-Advantage Filtering: We define the reward for a trajectory $\tau=(y_1, y_2)$ as being determined by the correctness of the final self-evaluation, i.e., $R(\tau) = C(y_2, y^)$. In our RL framework, the policy update relies on the estimated advantage $\hat{A}(\tau)$. For a mini-batch of $K$ trajectories $\{\tau_k\}_{k=1}^K$ sampled for a given query, if the reward same across all trajectories  the advantage estimate for each one becomes zero:
Consequently, the gradient of the primary policy objective vanishes, and the update is driven solely by the KL-divergence regularization term, $\nabla_\theta L(\theta) \approx -\beta \nabla_\theta D_{KL}(\pi_{\text{old}} \| \pi_\theta)$. To prevent such non-informative, potentially disruptive updates, we discard these zero-advantage mini-batches entirely.

Truncation Filtering: We discard any trajectory $\tau=(y_1, y_2)$ where either the initial response $y_1$ or the evaluation response $y_2$ is terminated prematurely by hitting the maximum sequence length. This prevents the model from being penalized with a poor reward for a potentially correct but verbose response, thereby maintaining the fidelity of the reward signal.



\subsection{A.3 Base Algorithm}

The core reinforcement learning algorithm we utilize is Group-wise Reward Policy Optimization (GRPO), an advanced on-policy method derived from the principles of Proximal Policy Optimization (PPO). GRPO is specifically designed to enhance training stability and efficiency by refining how rewards and policy updates are handled.

\textbf{Group-wise Advantage Estimation}
A key innovation in GRPO is its method for calculating the advantage function, which quantifies how much better a specific action is compared to the average action at a given state. Instead of relying on a single trajectory or a learned value function, GRPO operates on a group of samples.

For a given prompt $q$, the policy $\pi_{\theta_{old}}$ generates a batch of $G$ distinct output trajectories $\{o_1, o_2, \dots, o_G\}$. Each trajectory $o_i$ receives a corresponding reward $r_i$. The algorithm then computes the advantage $A_i$ for each trajectory by normalizing its reward against the statistics of the entire group. Specifically, the baseline is the mean reward of the group, and the result is scaled by the standard deviation. This is formally expressed as:

$$A_i = \frac{r_i - \operatorname{mean}(\{r_1, r_2, \dots, r_G\})}{\operatorname{std}(\{r_1, r_2, \dots, r_G\})}$$

This group-wise normalization makes the advantage estimate more robust and less susceptible to variance in reward scales, leading to more stable and reliable policy gradients during training.

\textbf{KL Divergence Regularization}
To prevent the policy from deviating too drastically from a trusted version during updates, GRPO incorporates a KL divergence penalty directly into its loss function. This term, $\mathbb{D}_{KL}(\pi_{\theta} || \pi_{ref})$, measures the difference between the current policy $\pi_{\theta}$ and a more stable reference policy $\pi_{ref}$ (which could be the policy from a previous iteration or a model from supervised fine-tuning). This acts as a soft constraint, ensuring that policy updates are kept within a reasonable trust region, which is crucial for maintaining training stability.

\textbf{The GRPO Objective Function}
The goal of GRPO is to maximize its objective function, which integrates the advantage estimate with the KL penalty. The optimization objective is defined as:


Here, $\beta$ is a coefficient that controls the strength of the KL regularization. The term $\hat{A}_i$ is a clipped version of the advantage, a technique inherited from PPO to further constrain policy updates:

$$\hat{A}_i = \min \left( \frac{\pi_{\theta}(o_i|q)}{\pi_{\theta_{old}}(o_i|q)} A_i, \operatorname{clip} \left( \frac{\pi_{\theta}(o_i|q)}{\pi_{\theta_{old}}(o_i|q)}, 1 - \varepsilon, 1 + \varepsilon \right) A_i \right)$$

The ratio $\frac{\pi_{\theta}(o_i|q)}{\pi_{\theta_{old}}(o_i|q)}$ compares the likelihood of an output under the new and old policies. By clipping this ratio within a small interval $[1-\varepsilon, 1+\varepsilon]$, the algorithm prevents excessively large updates that could destabilize the learning process.

\textbf{ Synergy with AdaPo and Computational Efficiency}
A significant practical reason for adopting GRPO is its computational synergy with our AdaPo framework. The batch of trajectories $\{o_1, o_2, \dots, o_G\}$ sampled for calculating the group-wise advantage in GRPO can be directly reused by AdaPo to estimate the training stage for each task. This dual-use of sampled data is highly efficient, as it completely eliminates the need for AdaPo to perform a separate, computationally expensive rollout sampling phase. This synergy streamlines the training pipeline, reduces overall computational overhead, and accelerates the learning process.




\subsection{A.4 Training Hyperparameter}
\begin{table}[!ht]
    \caption{Hyperparameters of Qwen2.5-VL-7B.}
    \label{tab:hyperparameters_and_training_schedule_large}
    \centering
    \setlength{\arrayrulewidth}{0.5pt}
 \resizebox{0.45\textwidth}{!}{
    \begin{tabular}{c|c|c}
    \toprule[2pt]
    \textbf{MLLM} & \textbf{Hyperparameter} & \textbf{Setting} \\ 
    \midrule
    \midrule
    \multirow{20}{*}{\makecell[c]{Qwen2.5-VL-7B}} & GPU & Tesla A100 (80GB) \\ 
     & \cellcolor{gray!15}Clip Ratio & \cellcolor{gray!15}0.25 \\ 
     & Top P & 0.99 \\ 
     & \cellcolor{gray!15}Rollout Repeat Size & \cellcolor{gray!15}8 \\
     & Temperature & 1.0 \\ 
     & \cellcolor{gray!15}Optimizer & \cellcolor{gray!15}Adam \\ 
     & KL Type & Low Var KL \\
     & \cellcolor{gray!15}Learning rate & \cellcolor{gray!15}5e-7 \\ 
        \multirow{8}{*}{} & Epochs & 2 \\
         & \cellcolor{gray!15}KL Coef $\beta_{base}$ & \cellcolor{gray!15}0.001 \\
    & KL Coef $\lambda$ & 0.01 \\ 
     & \cellcolor{gray!15}$R_{0\to1}$ & \cellcolor{gray!15} 1.0 \\
     & $R_{1\to1}$ & 1.0 \\ 
     & \cellcolor{gray!15}$R_{1\to0}$ & \cellcolor{gray!15}-1.0 \\ 
     & $R_{0\to0}$ & -1.0 \\
     & \cellcolor{gray!15} State Change $\theta$ & \cellcolor{gray!15}0.6 \\ 
        \multirow{8}{*}{} & Truncation Length & 5000 \\     
    & \cellcolor{gray!15} $K_{0\to1}$  & \cellcolor{gray!15} -1.0 \\ 
        \multirow{8}{*}{} & $K_{1\to1}$ & 1.0 \\   
        & \cellcolor{gray!15} $K_{1\to0}$  & \cellcolor{gray!15} 0.5 \\ 
        \multirow{8}{*}{} & $K_{0\to0}$ & -0.5 \\  
     \bottomrule[2pt]
    \end{tabular}
   }
    \vspace{-0.3cm}
\end{table}
\subsection{A.5 Hardware}
All experiments are performed on a CentOS Linux 7 server. The hardware specifications consist of 240GB of RAM, a 16-core Intel Xeon CPU, and eight NVIDIA A800 GPUs, each having 80GB of memory. 
\section{B Baselines}
The following section introduces the baseline methods used in our comparative experiments. These include Direct Supervised Fine-Tuning (SFT), Self-Taught Reasoner (STaR), Generalized Rejection-Sampling Policy Optimization (GRPO), and Self-Correction via Reinforcement Learning (SCoRe), covering supervised learning and reinforcement learning paradigms to comprehensively evaluate the effectiveness of our proposed AdaPO framework.

\textbf{Direct Supervised Fine-Tuning (SFT)}: As a standard baseline, we directly fine-tune the pre-trained model on the training datasets using a supervised learning paradigm. The fine-tuning process is conducted for 2 epochs with a batch size of 32. We employ the AdamW optimizer with a learning rate of $2 \times 10^{-5}$ and a linear learning rate scheduler with a warm-up ratio of 10\%. This setup ensures stable and effective learning on the downstream task.

\textbf{Self-Taught Reasoner (STaR)}: We implement the STaR methodology to enhance the model's reasoning capabilities. For each problem in the training set, the model first generates an initial response. Subsequently, it enters a self-correction loop for two rounds, where it attempts to refine and improve its previous answer. We then filter these generated responses, selecting only the correctly revised ones as new high quality training samples. This augmented dataset is used to fine-tune the model. The entire iterative process generation, refinement, and fine-tuning is repeated twice. The fine-tuning within each STaR iteration follows the same hyperparameter configuration as our SFT baseline, utilizing a learning rate of $2 \times 10^{-5}$, a batch size of 32, and the AdamW optimizer over 2 epochs.

\textbf{Generalized Rejection-Sampling Policy Optimization (GRPO)}: We fine-tune the model using the original GRPO algorithm on our consolidated dataset. In our implementation, we generate K=8 candidate responses for each prompt. The best response, as determined by our reward model, is selected for the policy update. The optimization is performed for 2 epochs using the AdamW optimizer with a learning rate of $2 \times 10^{-6}$. 

\textbf{ Self-Correction via Reinforcement Learning (SCoRe)}: We reproduce the two-stage reinforcement learning pipeline from SCoRe and apply it to fine-tune the model.
SCoRe (Self-Correction via Reinforcement Learning) is a two-stage multi-turn reinforcement learning approach designed to enhance large language models' intrinsic self-correction ability using self-generated data, addressing distribution shift and behavior collapse issues in existing methods. In Stage I, the goal is to obtain a policy initialization that improves second-attempt performance while constraining the first attempt to stay close to the base model (reference policy $\pi_{ref}$) to mitigate behavior collapse. The objective here is to maximize the reward of the second attempt (oracle reward for correctness) while enforcing a strict KL-divergence penalty on the first attempt, controlled by $\beta_2$, with the implementation using REINFORCE policy gradient with KL-divergence regularization to fine-tune the base model. Stage II focuses on multi-turn RL with reward shaping to jointly optimize both first and second attempts, incentivizing meaningful self-correction. Its objective is to maximize the sum of rewards, with a KL-divergence penalty (controlled by $\beta_1$) to align with the reference policy, and reward shaping involves a bonus term using $\alpha$ (where $\alpha$ > 1.0) to amplify transitions from incorrect to correct attempts and penalize regressions from correct to incorrect ones. For training hyperparameters,  with the optimizer being Adam i. The learning rate is $2 \times 10^{-5}$ . Batch sizes are 128, and the sampling temperature is 1.0. The reward shaping parameter $\alpha$ is set to 10 , the KL penalty $\beta_1$ is 0.01, and $\beta_2$ is 0.1.

\section{C Evaluation Benchmarks}
The following section presents detailed descriptions of the 8 benchmarks employed in our evaluation. These benchmarks span various domains, including mathematical reasoning across different visual contexts, multi-discipline multimodal understanding, and diagram interpretation, collectively providing a comprehensive assessment of the model's capabilities in complex visual-language reasoning tasks.

\textbf{Mathvista} is a benchmark designed to systematically evaluate the mathematical reasoning capabilities of foundation models in visual contexts. It comprises 6,141 examples, derived from 28 existing multimodal datasets involving mathematics and 3 newly created datasets (IQTest, FunctionQA, and PaperQA). Covering seven mathematical reasoning types (algebraic, arithmetic, geometry, logical, numeric common sense, scientific, and statistical reasoning) and five primary tasks (figure question answering, geometry problem solving, math word problem, textbook question answering, and visual question answering), it encompasses diverse visual contexts such as natural images, geometry diagrams, abstract scenes, and various charts/plots. 
 
\textbf{MathVerse} is a comprehensive and specialized visual math benchmark designed to equitably and in-depth evaluate the multi-modal mathematical reasoning capabilities of Multi-modal Large Language Models (MLLMs). It comprises 2,612 high-quality, multi-subject math problems with diagrams, curated from public sources and existing benchmarks, covering three primary areas: plane geometry, solid geometry, and functions, which are further divided into twelve fine-grained categories. Each problem is transformed by human annotators into six distinct versions with varying degrees of multi-modal information content (ranging from text-dominant to vision-only), resulting in a total of 15,672 test samples. 

\textbf{MATH-Vision} is a benchmark dataset designed to evaluate the multimodal mathematical reasoning capabilities of Large Multimodal Models (LMMs). It comprises 3,040 high-quality mathematical problems with visual contexts, carefully selected from 19 real math competitions spanning 12 grades. These problems are categorized into 16 distinct mathematical disciplines (e.g., algebra, analytic geometry, topology) and 5 difficulty levels (Level 1 being the easiest, Level 5 the most challenging). The dataset includes 1,532 multiple-choice questions and 1,508 open-ended questions, along with a smaller "testmini" subset of 304 questions for quick evaluation.

\textbf{WE-MATH} is a pioneering benchmark designed to explore the problem-solving principles of Large Multimodal Models (LMMs) in visual mathematical reasoning beyond end-to-end performance. It comprises 6.5K meticulously selected visual math problems, spanning 67 hierarchical knowledge concepts across 5 layers of knowledge granularity, covering areas like Plane Figures, Solid Figures, Transformations and Movements of Shapes, Positions and Directions, and Measurements.

\textbf{MMStar} is an elite vision-indispensable multi-modal benchmark comprising 1,500 challenging samples meticulously curated by humans, designed to accurately evaluate the multi-modal capabilities of Large Vision-Language Models (LVLMs). Addressing two key issues in existing benchmarks—unnecessary visual content in many samples and unintentional data leakage in LLM/LVLM training—MMStar ensures each sample is visually dependent, has minimal data leakage, and requires advanced multi-modal capabilities. 

\textbf{AI2D}(AI2 Diagrams) dataset is a newly compiled dataset for evaluating diagram interpretation tasks, created by researchers from the Allen Institute for Artificial Intelligence and the University of Washington. It comprises over 5,000 grade school science diagrams covering various topics, each annotated with exhaustive details including constituent segmentations, relationships between constituents, and their relationships to the diagram canvas. In total, AI2D contains annotations for more than 118,000 constituents and 53,000 relationships. 

\textbf{MMMU} (Massive Multi-discipline Multimodal Understanding and Reasoning Benchmark) is a benchmark designed to evaluate multimodal models on massive multi-discipline tasks requiring college-level subject knowledge and deliberate reasoning. It comprises 11.5K meticulously collected multimodal questions from college exams, quizzes, and textbooks, covering six core disciplines: Art \& Design, Business, Science, Health \& Medicine, Humanities \& Social Science, and Tech \& Engineering, spanning 30 subjects and 183 subfields. Featuring 30 highly heterogeneous image types such as charts, diagrams, maps, tables, music sheets, and chemical structures.

\textbf{MMMU-Pro} is a more robust version of the Massive Multi-discipline Multimodal Understanding and Reasoning (MMMU) benchmark, designed to rigorously assess multimodal models' true understanding and reasoning capabilities. It is constructed through a three-step process based on MMMU: filtering out questions answerable by text-only models, augmenting candidate options (from 4 to up to 10) to reduce guessing, and introducing a vision-only input setting where questions are embedded within screenshots or photos, challenging models to seamlessly integrate visual and textual information.
    \definecolor{mygreen}{RGB}{0,128,0}

\section{D Case Study}
\newtcolorbox{mybox}[1]{
    title={#1}, 
    colback=white, 
    colframe=gray!70, 
    fonttitle=\bfseries, 
    boxsep=5pt, 
    left=5pt,right=5pt,top=5pt,bottom=5pt 
}

\newtcolorbox{fullwidthbox}[1]{
    title={#1},
    colback=white, 
    colframe=gray!70, 
    fonttitle=\bfseries,
    boxsep=5pt,
    left=5pt,right=5pt,top=5pt,bottom=5pt,
    titlerule=0.5pt
}
\begin{figure*}[htbp] 
    \centering
    \includegraphics[width=0.2\textwidth]{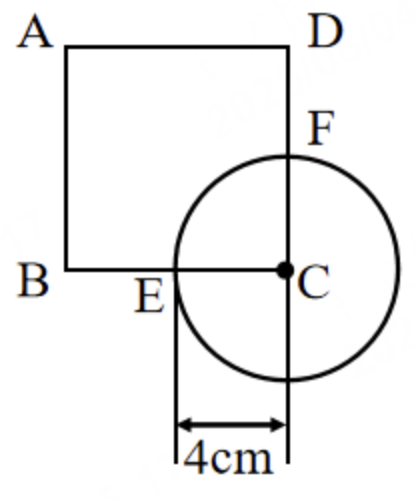} 
     \caption{Example 1}

\end{figure*}
\begin{figure*}[ht]
    \begin{fullwidthbox}{Prompt}
\textbf{Question:} As shown in the figure, a circle is drawn with vertex C of the square as the center. What is the measure of the central angle $\angle ECF$? ( )$^{\circ}$ A. 45; B. 60; C. 72; D. 90; E. No correct answer
A. 45; B. 60; C. 72; D. 90; E. No correct answer  
    \end{fullwidthbox}
\end{figure*}
\begin{figure*}[ht]
    \begin{fullwidthbox}{Model Response}
\textbf{ AdaPO turn 1 solution}: \textbf{\textcolor{mygreen}{(correct)}}

To determine the measure of the central angle $\angle ECF$, let's analyze the given figure step by step.

1.  Identify the properties of the square and the circle:
    The square has all sides of equal length.
    The circle is drawn with vertex C of the square as the center.
    The radius of the circle is the distance from the center C to any point on the circle. In this case, the radius is the side length of the square.
2.  Since the circle is drawn with vertex C of the square as the center, the radius of the circle is equal to the side length of the square. Let's denote the side length of the square as $s$. Therefore, the radius of the circle is $s$.
3.  In the figure, the radius of the circle is also the distance from C to E and from C to F. Since E and F are points on the circle, CE and CF are radii of the circle. Therefore, $CE = CF = s$.
4.  Since E and F are points on the circle and C is the center, $\angle ECF$ is an angle subtended by the arc EF at the center of the circle. In a square, the diagonals bisect each other at right angles. The diagonals of the square are the diameters of the circle.
5.  The diagonals of the square divide the square into four right-angled isosceles triangles. Each of these triangles has a right angle at the corner of the square and two 45-degree angles. Therefore, the diagonals of the square are equal in length and bisect each other at right angles.
6.  Since the diagonals of the square are diameters of the circle, the angle $\angle ECF$ is a right angle (90 degrees) because it is subtended by the diameter of the circle. Thus, the measure of the central angle $\angle ECF$ is 90 degrees.

The correct answer is: D. 90.

\textit{There might be an error in the solution, please evaluate the previous solution and provide a final answer.”}

\textbf{ AdaPO turn 2 solution}:\textbf{\textcolor{mygreen}{(correct)}} 

The previous reasoning process was correct in identifying the properties of the square and the circle, but let's summarize the steps again for clarity:
1.  Identify the properties of the square and the circle:
    - The square has all sides of equal length.
    - The circle is drawn with vertex C of the square as the center.
    - The radius of the circle is the distance from the center C to any point on the circle. In this case, the radius is the side length of the square.
2.  Since the circle is drawn with vertex C of the square as the center, the radius of the circle is equal to the side length of the square. Let's denote the side length of the square as $s$. Therefore, the radius of the circle is $s$.
3.  In the figure, the radius of the circle is also the distance from C to E and from C to F. Since E and F are points on the circle, CE and CF are radii of the circle. Therefore, $CE = CF = s$.
4.  Since E and F are points on the circle and C is the center, $\angle ECF$ is an angle subtended by the arc EF at the center of the circle. In a square, the diagonals bisect each other at right angles. The diagonals of the square are the diameters of the circle.
5.  The diagonals of the square divide the square into four right-angled isosceles triangles. Each of these triangles has a right angle at the corner of the square and two $45^{\circ}$ angles. Therefore, the diagonals of the square are equal in length and bisect each other at right angles.
6.  Since the diagonals of the square are diameters of the circle, the angle $\angle ECF$ is a right angle ($90^{\circ}$) because it is subtended by the diameter of the circle. Thus, the measure of the central angle $\angle ECF$ is $90^{\circ}$.

The correct answer is: D. 90
    \end{fullwidthbox}
\end{figure*}

\begin{figure*}[htbp] 
    \centering
    \includegraphics[width=0.3\textwidth]{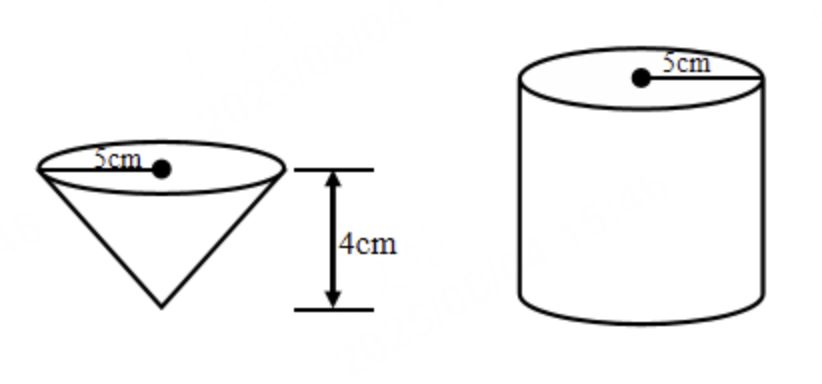} 
     \caption{Example 2}
\end{figure*}
\begin{figure*}[ht]
    \begin{fullwidthbox}{Prompt}
\textbf{Question:} The height of the cylindrical water storage container is as shown in the figure. Its height is equivalent to ( ) meters. A. 0.8; B. 0.08; C. 16; D. 8; E. No correct answer 
    \end{fullwidthbox}
\end{figure*}
\begin{figure*}[ht]
    \begin{fullwidthbox}{Model Response}
\textbf{ AdaPO turn 1 solution}: \textbf{\textcolor{mygreen}{(correct)}}
To determine the height of the cylindrical container, we need to compare the volumes of the conical and cylindrical containers. First, let's calculate the volume of the conical container. The formula for the volume of a cone is:
$$V_{\text{cone}} = \frac{1}{3} \pi r^2 h$$
where $r$ is the radius and $h$ is the height of the cone. From the figure, we can see that the radius $r$ of the conical container is 3 cm and the height $h$ is 4 cm. Substituting these values into the formula, we get:
$$V_{\text{cone}} = \frac{1}{3} \pi (3)^2 (4) = \frac{1}{3} \pi (9) (4) = \frac{1}{3} \pi (36) = 12 \pi \text{ cm}^3$$
Next, let's calculate the volume of the cylindrical container. The formula for the volume of a cylinder is:
$$V_{\text{cylinder}} = \pi r^2 h$$
where $r$ is the radius and $h$ is the height of the cylinder. From the figure, we can see that the radius $r$ of the cylindrical container is 5 cm. Let the height of the cylindrical container be $H$ cm. Substituting these values into the formula, we get:
$$V_{\text{cylinder}} = \pi (5)^2 H = 25 \pi H \text{ cm}^3$$
According to the problem, after pouring water 6 times from the conical container into the cylindrical container, the cylindrical container is just full. Therefore, the total volume of water poured from the conical container is equal to the volume of the cylindrical container:
$$6 \times V_{\text{cone}} = V_{\text{cylinder}}$$
Substituting the volumes we calculated:
$$6 \times 12 \pi = 25 \pi H$$
Simplifying this equation:
$$72 \pi = 25 \pi H$$
Dividing both sides by $25 \pi$:
$$H = \frac{72 \pi}{25 \pi} = \frac{72}{25} = 2.88 \text{ cm}$$
Since the height of the cylindrical container is 2.88 cm, we need to convert this to meters:
$$2.88 \text{ cm} = 0.0288 \text{ m}$$
However, the closest option to 0.0288 m is 0.08 m. Therefore, the correct answer is: $\boxed{B}$

    \end{fullwidthbox}
\end{figure*}
\begin{figure*}[ht]
    \begin{fullwidthbox}{Model Response}
\textit{There might be an error in the solution, please evaluate the previous solution and provide a final answer.”}

\textbf{ AdaPO turn 2 solution}:\textbf{\textcolor{mygreen}{(correct)}} 

The previous reasoning process was correct in its approach but made a small mistake in the final step of converting the height from centimeters to meters. Let's go through it again with a detailed reflection on the steps and the final answer.

1.  Calculate the volume of the conical container: The formula for the volume of a cone is:
    $$V_{\text{cone}} = \frac{1}{3} \pi r^2 h$$
    From the figure, the radius $r$ of the conical container is 3 cm and the height $h$ is 4 cm. Substituting these values into the formula, we get:
    $$V_{\text{cone}} = \frac{1}{3} \pi (3)^2 (4) = \frac{1}{3} \pi (9) (4) = \frac{1}{3} \pi (36) = 12 \pi \text{ cm}^3$$

2.  Calculate the volume of the cylindrical container: The formula for the volume of a cylinder is:
    $$V_{\text{cylinder}} = \pi r^2 h$$
    From the figure, the radius $r$ of the cylindrical container is 5 cm. Let the height of the cylindrical container be $H$ cm. Substituting these values into the formula, we get:
    $$V_{\text{cylinder}} = \pi (5)^2 H = 25 \pi H \text{ cm}^3$$

3.  Set up the equation based on the problem statement: According to the problem, after pouring water 6 times from the conical container into the cylindrical container, the cylindrical container is just full. Therefore, the total volume of water poured from the conical container is equal to the volume of the cylindrical container:
    $$6 \times V_{\text{cone}} = V_{\text{cylinder}}$$
    Substituting the volumes we calculated:
    $$6 \times 12 \pi = 25 \pi H$$
    Simplifying this equation:
    $$72 \pi = 25 \pi H$$
    Dividing both sides by $25 \pi$:
    $$H = \frac{72 \pi}{25 \pi} = \frac{72}{25} = 2.88 \text{ cm}$$

4.  Convert the height from centimeters to meters: Since the height of the cylindrical container is 2.88 cm, we need to convert this to meters:
    $$2.88 \text{ cm} = 0.0288 \text{ m}$$
    
Conclusion

The closest option to 0.0288 m is 0.08 m. Therefore, the correct answer is:
$$\boxed{B}$$
    \end{fullwidthbox}
\end{figure*}

\begin{figure*}[htbp] 
    \centering
    \includegraphics[width=0.2\textwidth]{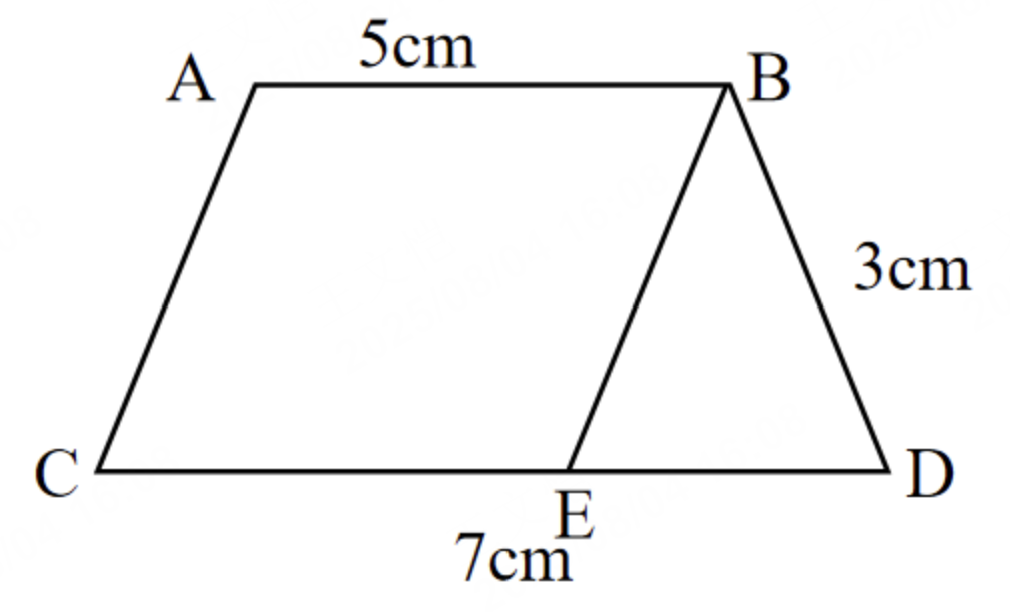} 
     \caption{Example 3}
\end{figure*}
\begin{figure*}[ht]
    \begin{fullwidthbox}{Prompt}
\textbf{Question:} As shown in the figure, an isosceles trapezoid is divided into a parallelogram and a triangle. The perimeter of the triangle is () centimeters. A. 24; B. 20; C. 18; D. 8; E. No correct answer
    \end{fullwidthbox}
\end{figure*}

\begin{figure*}[ht]
    \begin{fullwidthbox}{Model Response}
\textbf{ AdaPO turn 1 solution}: \textbf{\textcolor{red}{(incorrect)}}

To solve the problem, let's break down the information given and the geometry of the figure. We have an isosceles trapezoid $ABCD$ with $AB$ and $CD$ as the parallel sides. The lengths of the sides are:
- $AB = 5 \text{ cm}$
- $CD = 7 \text{ cm}$
- $BC = 3 \text{ cm}$

The trapezoid is divided into a parallelogram and a triangle. The parallelogram is $ABCE$ and the triangle is $BDE$. Since $ABCD$ is an isosceles trapezoid, the non-parallel sides $AD$ and $BC$ are equal in length. However, we are given the length of $BC$ as 3 cm, and we need to find the length of $DE$.

In the triangle $BDE$, we know:
- $BD = 3 \text{ cm}$
- $DE = 7 - 5 = 2 \text{ cm}$ (since $CD = 7 \text{ cm}$ and $AB = 5 \text{ cm}$)

Now, we need to find the perimeter of the triangle $BDE$. The perimeter of a triangle is the sum of the lengths of its sides. Therefore, the perimeter of triangle $BDE$ is:
$$BD + DE + BE$$

From the problem, we know:
- $BD = 3 \text{ cm}$
- $DE = 2 \text{ cm}$
- $BE$ is the remaining part of the trapezoid, which is the same as the length of $CE$ (since $ABCE$ is a parallelogram and $AB = CE = 5 \text{ cm}$)

So, the perimeter of the triangle $BDE$ is:
$$BD + DE + BE = 3 + 2 + 5 = 10 \text{ cm}$$

However, the problem states that the perimeter of the triangle is one of the given options. Let's re-evaluate the options:
A. 24
B. 20
C. 18
D. 8
E. No correct answer

The correct perimeter of the triangle $BDE$ is 10 cm, but since the options do not include 10 cm, the correct answer is:
E. No correct answer

Thus, the answer is: $\boxed{E}$

\textit{There might be an error in the solution, please evaluate the previous solution and provide a final answer.”}

\textbf{ AdaPO turn 2 solution}:\textbf{\textcolor{mygreen}{(correct)}} 

Let's re-evaluate the problem step by step. We have an isosceles trapezoid $ABCD$ with $AB$ and $CD$ as the parallel sides. The lengths of the sides are:
$AB = 5 \text{ cm}$
$CD = 7 \text{ cm}$
$BC = 3 \text{ cm}$

The trapezoid is divided into a parallelogram $ABCE$ and a triangle $BDE$. Since $ABCD$ is an isosceles trapezoid, the non-parallel sides $AD$ and $BC$ are equal in length. However, we are given the length of $BC$ as 3 cm, and we need to find the length of $DE$.

In the triangle $BDE$, we know:
$BD = 3 \text{ cm}$
$DE = 7 - 5 = 2 \text{ cm}$ (since $CD = 7 \text{ cm}$ and $AB = 5 \text{ cm}$)

Now, we need to find the length of $BE$. Since $ABCE$ is a parallelogram, $AB = CE = 5 \text{ cm}$ and $AE = BC = 3 \text{ cm}$. Therefore, $BE = AE = 3 \text{ cm}$.

Now, we can find the perimeter of the triangle $BDE$. The perimeter of a triangle is the sum of the lengths of its sides. Therefore, the perimeter of triangle $BDE$ is:
$$BD + DE + BE = 3 + 2 + 3 = 8 \text{ cm}$$
So, the correct perimeter of the triangle $BDE$ is 8 cm.

Thus, the answer is: $\boxed{D}$
    \end{fullwidthbox}
\end{figure*}

\begin{figure*}[htbp] 
    \centering
    \includegraphics[width=0.4\textwidth]{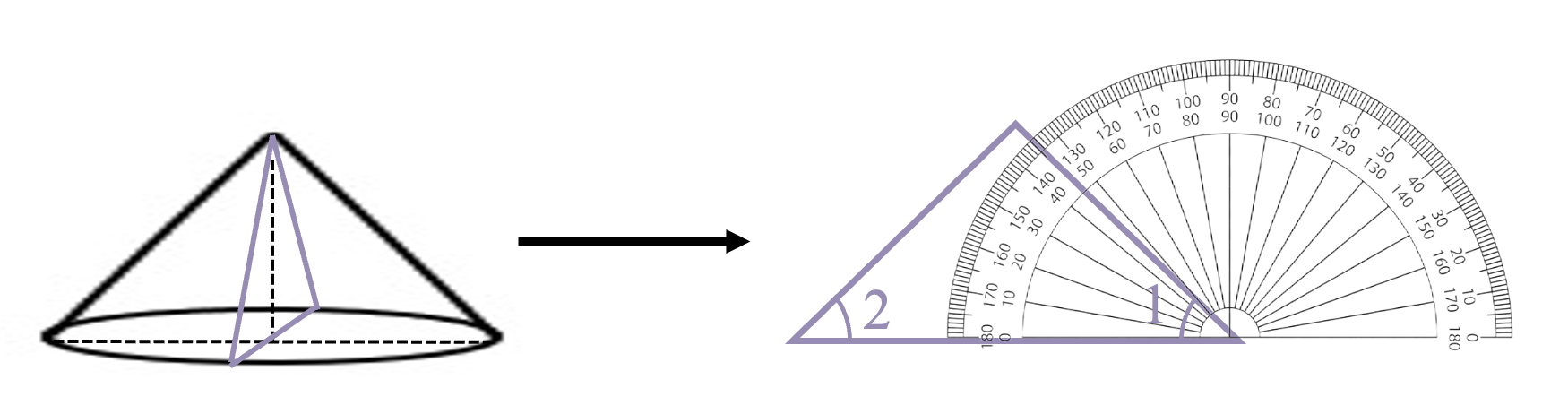} 
     \caption{Example 4}
\end{figure*}
\begin{figure*}[ht]
    \begin{fullwidthbox}{Prompt}
\textbf{Question:} As shown in the diagram, triangle ABC is a cross-section of a cone. What is the measure of $\angle 2$? A. $30^{\circ}$; B. $45^{\circ}$; C. $60^{\circ}$; D. $90^{\circ}$; E. No correct answer
    \end{fullwidthbox}
\end{figure*}

\begin{figure*}[ht]
    \begin{fullwidthbox}{Model Response}
\textbf{ AdaPO turn 1 solution}: \textbf{\textcolor{red}{(incorrect)}}

To determine the measure of $\angle 2$, let's analyze the given information step by step.

1.  The diagram shows a cone with a cross-section that is an isosceles triangle. The base of this triangle is the diameter of the base of the cone, and the two equal sides are the slant heights of the cone.
2.  When we look at the isosceles triangle formed by the cross-section, we can see that it is also a right triangle because the line from the apex of the cone to the midpoint of the base (which is the radius of the base of the cone) forms a right angle with the radius of the base.
3.  In the right triangle, the angle opposite the radius (which is half the diameter) is $30^{\circ}$ because the triangle is isosceles and the angles in a triangle sum up to $180^{\circ}$. Therefore, the two base angles of the isosceles triangle are equal, and since one of them is $30^{\circ}$, the other must also be $30^{\circ}$.
4.  The angle $\angle 2$ is the angle opposite the radius in the right triangle, which is $30^{\circ}$. Thus, the measure of $\angle 2$ is $30^{\circ}$.

The correct answer is: A. $30^{\circ}$

\textit{There might be an error in the solution, please evaluate the previous solution and provide a final answer.”}

\textbf{ AdaPO turn 2 solution}:\textbf{\textcolor{mygreen}{(correct)}} 

The previous reasoning process was correct in identifying that the triangle formed by the cross-section of the cone is an isosceles triangle and that it is also a right triangle. However, the final step of identifying the measure of $\angle 2$ was a bit rushed. Let's go through it again in detail.

1.  The cross-section of the cone is an isosceles triangle with the base as the diameter of the base of the cone and the two equal sides as the slant heights of the cone.
2.  In this isosceles triangle, the line from the apex of the cone to the midpoint of the base (which is the radius of the base of the cone) forms a right angle with the radius of the base of the cone. This is because the radius of the base of the cone is perpendicular to the tangent at the point of tangency, and the line from the apex to the midpoint of the base is a radius of the cone's lateral surface.
3.  Therefore, the triangle formed is a right triangle with one angle of $90^{\circ}$. Since it is an isosceles triangle, the two base angles are equal. Let's denote the angles as follows:
    The right angle is $90^{\circ}$.
    The two base angles are equal, and let's call each of them $\angle 1$ and $\angle 2$.
    Since the sum of the angles in a triangle is $180^{\circ}$, we have:
        $$90^{\circ} + \angle 1 + \angle 2 = 180^{\circ}$$
    Because the triangle is isosceles, $\angle 1 = \angle 2$. So, we can write:
        $$90^{\circ} + \angle 2 + \angle 2 = 180^{\circ}$$       $$90^{\circ} + 2\angle 2 = 180^{\circ}$$       $$2\angle 2 = 90^{\circ}$$       $$\angle 2 = 45^{\circ}$$

Thus, the measure of $\angle 2$ is $45^{\circ}$.

The correct answer is: B. $45^{\circ}$
    \end{fullwidthbox}
\end{figure*}

\end{document}